\ifdefined\pdfminorversion\pdfminorversion=7\fi
\documentclass{gpl_preprint}

\usepackage{amsmath,amssymb,amsfonts}
\usepackage{array}
\usepackage[capitalize]{cleveref}  

\usepackage{amsmath,amsfonts,bm}

\def\eqref#1{equation~\ref{#1}}

\def\1{\bm{1}}

\DeclareMathAlphabet{\mathsfit}{\encodingdefault}{\sfdefault}{m}{sl}
\SetMathAlphabet{\mathsfit}{bold}{\encodingdefault}{\sfdefault}{bx}{n}

\definecolor{notecol}{HTML}{2c7fb8}   
\definecolor{srccol}{HTML}{7a7a7a}    
\definecolor{todocol}{HTML}{c0392b}   

\newcommand{\onote}[1]{}

  {\begin{itemize}\setlength{\itemsep}{2pt}\setlength{\parskip}{0pt}\setlength{\topsep}{2pt}}%
  {\end{itemize}}

\newtheorem{proposition}{Proposition}
\newtheorem{corollary}{Corollary}

\newcommand{\zr}{z}                       
\newcommand{\Ccond}{c}                     
\newcommand{\zdet}{z_{\text{det}}}         

\newcommand{\pizero}{$\pi_0$}
\newcommand{\pizerofive}{$\pi_{0.5}$}
\newcommand{\GPL}{\textsc{gpl}}            
\newcommand{\SR}{\text{SR}}

\renewcommand{\GPL}{{\rmfamily\textsc{gpl}}}

\graphicspath{{figures/}{../figures/}}

\titleformat*{\paragraph}{\normalfont\sffamily\bfseries}

\title{Reasoning as a Distribution: Generative Plan Latents for Vision--Language--Action Models}
\renewcommand{\authorlist}{%
  \authorformat[1]{Matteo Cercola}, \authorformat[2]{Katie Z Luo},
  \authorformat[2]{Milan Ganai}\\[3pt]
  \authorformat[2,3]{John Irvin Alora}, \authorformat[1]{Simone Formentin},
  \authorformat[4,2]{Marco Pavone}%
}
\affiliation[1]{Politecnico di Milano}
\affiliation[2]{Stanford University}
\affiliation[3]{DAF AI Studio}
\affiliation[4]{NVIDIA}
\metadata[Project page]{\url{https://matcercola18.github.io/reasoning-as-a-distribution/}}
\hypersetup{
  pdftitle={Reasoning as a Distribution: Generative Plan Latents for Vision-Language-Action Models},
  pdfauthor={Matteo Cercola, Katie Z Luo, Milan Ganai, John Irvin Alora, Simone Formentin, Marco Pavone},
  pdfsubject={Research preprint}
}
\abstracttext{\begin{abstract}
Vision–Language–Action (VLA) policies increasingly reason before acting, but existing reasoning models compress this intermediate decision into a single representation. When several plans are valid, this commitment hides alternatives that could otherwise be selected or revised. We introduce the Generative Plan Latent (GPL), which represents the reasoning state as a low-dimensional conditional distribution over plans, such as which object, grasp, or route to use. This makes the reasoning state \emph{addressable}: without retraining the VLA, plans can be sampled (\emph{coverage}), selected on request (\emph{steering}), and reweighted by deployment outcomes (\emph{outcome adaptation}). We test all three on SimplerEnv manipulation with $\pi_0$, and extend the analysis to RoboCasa, LIBERO, and a maze domain. Steering the plan raises the fraction of executions that follow the request from 0.48 to 0.94, while the same guidance on the native action distribution shows no reliable gain. Finally, we show that an explicit plan is most useful when it carries decision information not already determined by the observation.
\end{abstract}
}

\begin{document}
\maketitle

\section{Introduction}
\label{sec:introduction}

Vision–Language–Action (VLA) policies increasingly place a reasoning stage
between perception and action
\citep{Zawalski24-ecot,hao2024training,bai2026latent,
huang2026thinkact,huang2026fast}, and this intermediate state empirically
improves policy success rate. This intermediate state is attractive because it can separate what the robot decides to do from the motor trajectory that realizes that decision. Yet existing reasoning models represent this state as one vector per observation. This raises the question: should a model expose one committed decision per observation, or
a distribution over valid ones?

\begin{figure}[t]
    \centering
    \includegraphics[width=0.68\linewidth]{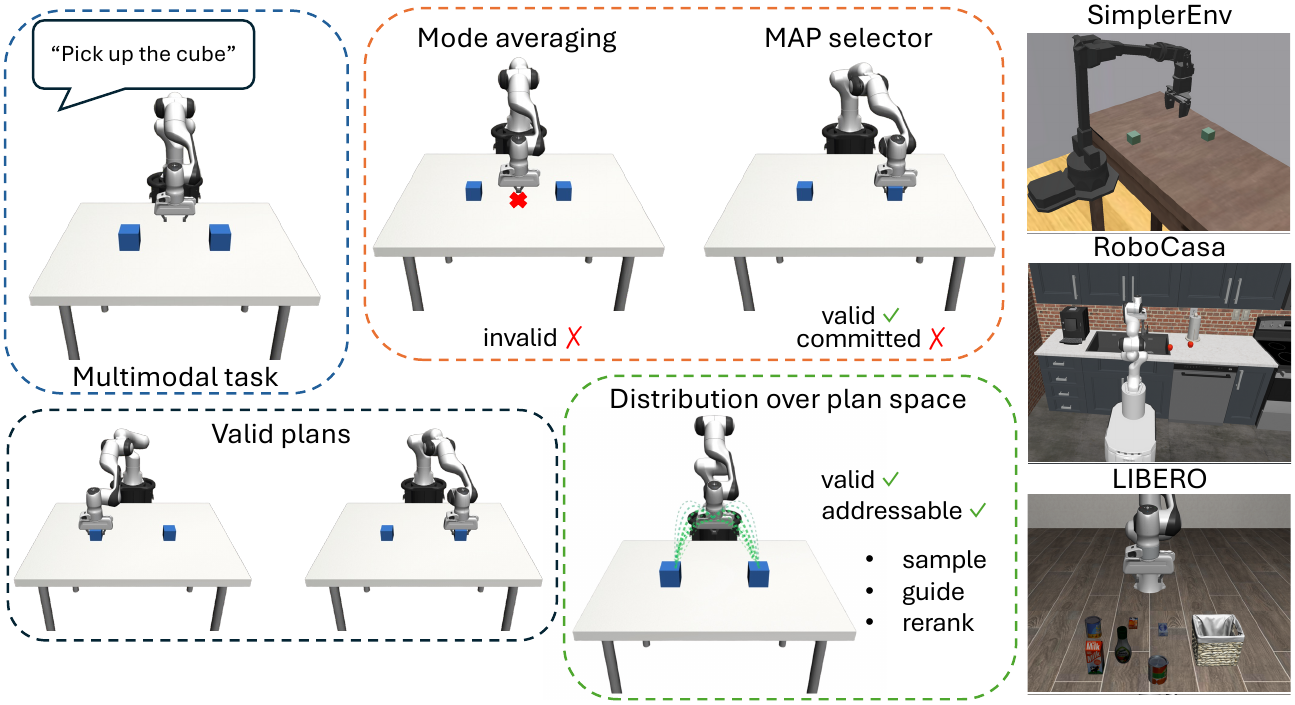}
    \caption{
    \textbf{The distribution should live at the decision bottleneck.}
    When several plans are valid, a committed regression readout can lie
    between them; \GPL{} preserves the alternatives at the plan level, where
    they can be sampled, guided, or reweighted before execution.
    }
    \label{fig:motivation}
\end{figure}

The distinction matters whenever the observation leaves more than one decision valid. Consider a robot facing two identical cubes, with an instruction that names neither: nothing favors one target over the other. The same ambiguity is pervasive in everyday tasks: whether to pick up a can or push it aside, how to grasp a mug, or which route to take. A reasoning state that must produce a single decision necessarily collapses these alternatives. Averaging them may yield an invalid plan between the cubes; choosing one avoids this failure but still hides the equally valid alternative. The missing property is therefore not simply validity, but a representation that preserves the alternatives so they remain accessible at inference time, allowing preferences or outcomes to dynamically select the most viable strategy at deployment without retraining (\Cref{fig:motivation}). We call a reasoning state \emph{addressable} when, at inference time, the policy can be: sampled to obtain distinct valid decisions (\emph{coverage}), conditioned to select among them a single decision (\emph{steering}), and reweighted by outcome to adapt the decision in closed-loop (\emph{outcome adaptation}).
The limitation of prior
designs is therefore a lack of addressability. 
This is \emph{not} the usual action-multimodality problem: robot learning has
already developed generative action models precisely to avoid averaging
incompatible motor trajectories (\Cref{sec:related}).
The unresolved question is whether the \emph{upstream decision}
should itself also be represented as a distribution. 

We introduce the \emph{Generative Plan Latent} (\GPL{}), which replaces a
deterministic plan readout with a conditional distribution \(p_\theta(z\mid c)\) of plans $z$ given a VLA representation $c$. This distribution can be sampled, steered toward a requested mode, and reweighted from
deployment experience, all without retraining the underlying VLA. We show that the plan provides a substantially stronger handle than the action distribution for expressing an external preference. Under matched executors, selecting a requested plan increases the fraction of executions that follow it from 0.48 to 0.94, whereas applying the same inference-time intervention to \pizero{}'s native action distribution yields no reliable improvement. Because these alternatives remain explicit, deployment outcomes can also change which plans are preferred without updating the plan generator or VLA. This benefit is conditional: an explicit plan can influence behavior only when it carries decision information that the executor cannot already recover from the observation. We call this the \emph{redundancy principle}. In a controlled test, changing only the demonstrations to reduce this redundancy increases the plan-selection effect from +0.19 to +0.65. Finally, these capabilities do not depend on hand-designed plan coordinates: they persist when the plan representation is learned directly from demonstrated behavior.

Our contributions are:
\begin{itemize}
    \item We identify the decision bottleneck as a distinct locus for multimodality in VLAs and define \emph{addressability}, the capacity to preserve alternative decisions so they can be sampled, selected, and revised. Matched plan- and action-level interventions show that the decision bottleneck provides the stronger handle for controlling behavior.

    \item We introduce \GPL{}, a distribution over task-level plans,
    instantiated by a rectified-flow head and integrated into \pizero{} with
    frozen base weights and a small adaptation.

    \item We propose the \emph{redundancy principle}: an explicit plan should affect behavior only when it carries decision information not already determined by the observation.
\end{itemize}
\section{Related Work}
\label{sec:related}

\paragraph{Reasoning in VLAs.}
Recent VLAs increasingly insert an intermediate reasoning stage between
perception and action. Embodied Chain-of-Thought (ECoT) \citep{Zawalski24-ecot} makes this reasoning
explicit in language, grounded in robot-relevant quantities such as subtasks,
objects, and motion; other approaches move this computation into continuous
hidden states
\citep{huang2026thinkact,huang2026fast,bai2026latent,tur2026rdvla}. CoT-VLA \citep{zhao2025cot} instead samples visual subgoals as
sequences of future frames, yielding a visual chain-of-thought in pixel space
rather than a compact plan representation. Despite these different representations, continuous action-domain reasoners typically expose a single intermediate state per observation (\Cref{tab:cell}), rather than an explicit distribution over alternative
decisions.

\paragraph{Continuous and generative latent reasoning.}
Continuous reasoning was first studied in the language domain: Coconut
\citep{hao2024training} feeds hidden states back into the model as continuous
thoughts, while generative approaches place a distribution over the reasoning
state
\citep{ye2024diffusion,huang2025reinforcing,kang2026ladirlatentdiffusionenhances,tu2026latent}.
STAR-LDM \citep{lovelace2025stopthinkautoregress} similarly samples a latent reasoning state
that is injected as soft tokens before generation. These works establish
continuous and stochastic reasoning for text.

\paragraph{Latent plans and generative planning.}
Sampling latent plans to condition low-level control is well established in
robot learning
\citep{lynch2019learning,lynch2021language,mees2022hulc}.
Higher-level policies can search these spaces
\citep{rosete2022tacorl}, while LDCQ
\citep{venkatraman2024reasoning} learns a multimodal distribution over latent
skills and selects among them with a value function.
Trajectory-level generative planners
\citep{janner2022planning,ajay2023is} combine sampling with test-time
guidance, while SuSIE \citep{black2024zero} and UniPi
\citep{du2023learning} generate visual subgoals for downstream control.

\paragraph{Generative actions and adaptation.}
Generative action models
\citep{florence2022implicit,lee2024behavior,chi2025diffusion} and flow-based
VLAs \citep{black2024pi_0} already represent multimodal motor behavior.
Flow-reversal and noise-space methods
\citep{wagenmaker2025steering,tang2026improving,lu2026unisteerunifiednoisesteering} steer these
action distributions, while verifier-based selection
\citep{kwok2025robomonkeyscalingtesttimesampling}, rejection-based post-training
\citep{tan2025interactive}, reinforcement-learning-style updates
\citep{chen2025conrft,li2025simplevla}, and failure- or energy-guided action
generation \citep{zheng2026failing,song2026omniguide} use feedback to improve
behavior.

Together, these lines of work provide intermediate reasoning states,
generative latent plans, and multimodal action distributions, but leave open
where multimodality should reside in a reasoning-augmented VLA.
\section{Method: The Generative Plan Latent}
\label{sec:method}

\begin{figure}[t]
    \centering
    \includegraphics[width=0.66\linewidth]{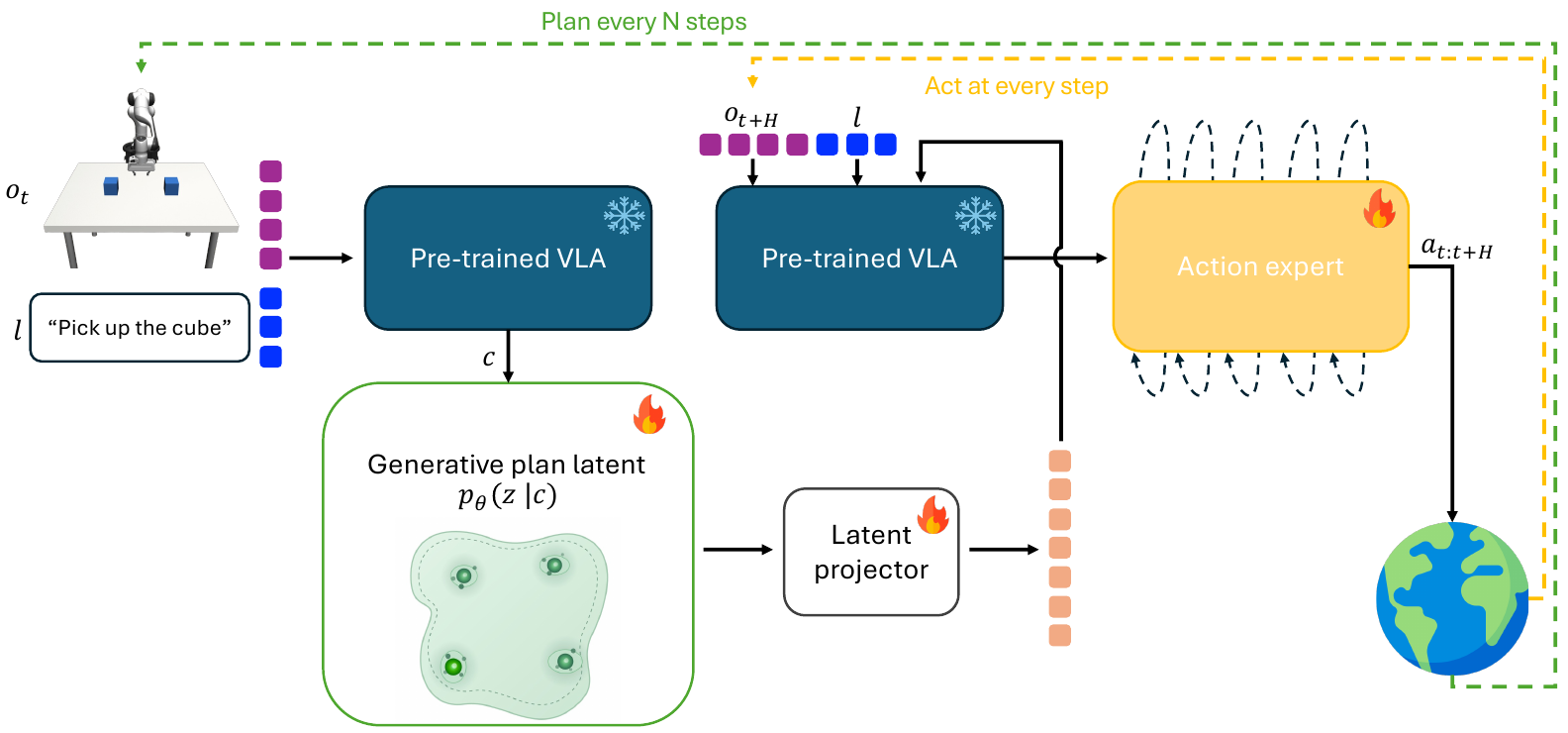}
    \caption{
        \textbf{Generative Plan Latent (\GPL{}).}
        The frozen VLA backbone encodes $(o,l)$ into $c$; a small
        rectified-flow head samples a plan $z \sim p_\theta(z\mid c)$; a
        projector injects $z$ as $k=8$ soft tokens into the executor prefix.
        Sampling, guidance, and outcome-based reranking all operate on this
        one plan distribution.
    }
    \label{fig:architecture}
\end{figure}

\paragraph{The change relative to a standard VLA.}
A modern VLA may already be generative at the action level; our method does
not replace that machinery. Instead, we insert a generative variable
\emph{upstream} of the action model, at the point where the task-level
decision is represented. \Cref{fig:architecture} traces the resulting
pipeline: the frozen VLA backbone reads the observation and instruction into a
conditioning vector, the generative plan
latent samples and projects the task-level decision from that conditioning,
and the action expert consumes the projected plan alongside fresh observations
to act at every step.

\paragraph{Problem setting.}
Let $o$ denote an image observation and $l$ a language instruction.
A pretrained VLA backbone $f_\phi$ maps them to a conditioning representation
$c = f_\phi(o,l)$.
We define a plan $z\in\mathbb{R}^{d_z}$ as a low-dimensional representation of
the \emph{decision} required by the task as opposed to the motor trajectory that
accomplishes it.

\paragraph{Generative plan model.}
We model the conditional plan distribution $p(z\mid c)$ with rectified flow
\citep{liu2023flow,lipman2023flow}.
With $z_0\sim\mathcal{N}(0,I)$, $t\sim\mathcal{U}[0,1]$ and
$z_t=(1-t)z_0+tz$ for a plan $z$ extracted from a demonstration, a velocity
field $v_\theta$ is trained with
\begin{equation}
    \mathcal{L}_{\mathrm{plan}}(\theta) = \mathbb{E}_{c,z,z_0,t} \left\| v_\theta(z_t,t,c) - (z-z_0) \right\|_2^2.
\end{equation}
At inference we draw $z_0$ from the prior and integrate with the Euler method
to obtain $z \sim p_\theta(z\mid c)$; because this takes only a handful of
integration steps rather than a token-by-token decode, one latent-plan draw
costs roughly \(80\times\) less than one sampled reasoning chain in a matched
comparison on the same conditioning and backbone (ECoT-VLA, 
\cite{Zawalski24-ecot}).
A projector $P_\eta$ maps the sampled plan to $k=8$ soft tokens
$[e_1,\ldots,e_k]$ inserted into the executor prefix, the \emph{plan interface}, after which the executor
predicts its action chunk using its native action objective.

\paragraph{Steering.}
Suppose a user specifies a desired plan property $y$, such as which object to
grasp or which route to follow. We train a lightweight \emph{noise-conditioned}
classifier $q_\psi(y\mid z_t,t,c)$ on intermediate noisy plans along the flow
path and bias the velocity field with its score during inference:
\begin{equation}
    v_{\mathrm{guide}} (z_t,t,c;y) = v_\theta(z_t,t,c) + w\, \nabla_{z_t} \log q_\psi(y\mid z_t,t,c),
    \label{eq:plan_guidance}
\end{equation}
where $w\geq0$ controls guidance strength.
Setting $w=0$ recovers the original plan distribution while increasing $w$ moves
probability mass toward the requested mode, and neither the plan generator
nor the VLA is retrained when the command changes (\Cref{subsec:steer}).

\paragraph{Outcome adaptation from experience.}
Steering assumes that the desired plan property is specified externally;
outcome adaptation instead uses feedback obtained after deployment, while
keeping the plan generator fixed. Let
\(z^{(1)},\ldots,z^{(K)}\sim p_\theta(z\mid c)\) be candidate plans and let
\(s(c,z)\) denote a deployment-time score, with larger values preferred.
Outcome adaptation reweights the plan prior toward high-scoring candidates:
\begin{equation}
    p_{\mathrm{adapt}}(z\mid c)
    \propto
    p_\theta(z\mid c)
    \exp\!\left(\beta s(c,z)\right),
    \label{eq:si_distribution}
\end{equation}
where \(\beta\geq0\) controls how strongly deployment feedback modifies the
plan prior. The score
\(s\) is deliberately generic, and we evaluate two broad classes of scoring mechanisms (formalized in \Cref{app:si_details}). A \emph{learned critic} trains a parametric model \(s(c,z)=g_\omega(z)\) to predict rewards directly from accumulated \((z,r)\) deployment rollouts, where \(r\) is the deployment outcome associated with \(z\). Alternatively, when a task partitions into discrete modes, a non-parametric \emph{reward table} simply assigns a candidate plan the historical average reward of its corresponding mode, requiring zero training floating-point operations (FLOPs). Rather than greedily executing a single top-scoring draw, both methods reweight the discrete candidate pool with Boltzmann weights; reported performance is the resulting expected outcome. Both scoring mechanisms can continuously absorb new deployment experience to update their estimates, while the generative prior \(p_\theta(z\mid c)\) and the underlying VLA remain completely fixed.

\paragraph{Adapting a pretrained VLA.}
While the GPL is designed to augment any modern pretrained VLA, we instantiate our primary architecture using \pizero{}. To integrate the generative plan without disrupting the underlying capabilities of the backbone, we add a projector and a low-rank adapter (LoRA) \citep{hu2021lora} to the action expert. This adapter is trained purely through the native action objective of the VLA without requiring an auxiliary plan-imitation loss. To ensure the executor remains robust, we use classifier-free guidance (CFG) \citep{ho2022classifierfreediffusionguidance}, so we randomly drop plan tokens during training. We evaluate against three strict baselines: the unmodified stock \pizero{}, a plan-less control (which uses our adapted weights but omits the plan at inference), and a dedicated no-plan adapter trained at an identical budget. Full details are in \Cref{app:pi0}.
\section{Three Capabilities of an Addressable Plan}
\label{sec:experiments}

\paragraph{Setup.}
We deploy our method across two distinct executor pipelines. To isolate reasoning from execution confounds and enable extensive ablation matrices, we first use a learned soft-token executor featuring a frozen perception backbone and a flow action head trained from scratch. For external validity, we then evaluate on a pretrained VLA (\pizero{} or $\pi_{0.5}$), adapted as described in \Cref{sec:method}.


\subsection{Coverage: Representing the Full Decision Space}
\label{subsec:coverage}

Coverage asks whether the reasoning state preserves all valid decisions rather
than merely returning one valid point. We therefore compare GPL not only with
a deterministic mean-squared-error (MSE) readout, which can average incompatible modes, but also
with a winner-take-all (WTA) selector that commits to the most likely valid
mode without maintaining a distribution. We test this on a square-footprint object that admits two valid top-down orientations $90^\circ$ apart. As detailed in \Cref{tab:addressable-main}, the deterministic MSE baseline averages the two modes into a diagonal grasp that misses the object entirely. While WTA nearly eliminates invalid predictions, it collapses to a single mode and fails to cover all available decisions. GPL, instead, successfully captures both valid modes across all layouts. This shows that avoiding an invalid mean is insufficient and that true coverage requires preserving the alternative decisions.

\begin{table}[t]
\centering
\caption{\textbf{Avoiding the mean is insufficient.} Eight held-out layouts,
five seeds, \(K=16\). The winner-take-all selector removes mode averaging
almost completely but still sits at the floor on coverage and steerability.}
\label{tab:addressable-main}
\footnotesize 
\setlength{\tabcolsep}{3.5pt}
\begin{tabular}{lccccc}
\toprule
arm & invalid draw & P(cover all) & steer range & rerank feasible & addressable \\
\midrule
deterministic MSE twin & 1.00 & 0.00 & 0.00 & 0.00 & no \\
deterministic WTA (MAP) & 0.03 & 0.00 & 0.00 & 0.50 & no \\
\GPL{} (flow) & 0.14 & 1.00 & +0.31 & 1.00 & yes \\
\bottomrule
\end{tabular}
\end{table}


\subsection{Steerability: The Plan Is the Stronger Handle}
\label{subsec:steer}

While coverage exposes alternatives, steerability asks whether a user can choose
among them without retraining the base models, and whether the 
\emph{locus} of that guidance matters. We report the
\emph{commanded-mode rate}, the fraction of episodes whose executed decision
matches the request, and the \emph{steer range}, i.e., difference between
opposite commands.

\paragraph{The plan distribution can be steered.}
On the two-object task, where training demonstrations are perfectly balanced between the two modes, the requested mode is executed with probability $0.41$ without guidance. By using a lightweight noise-conditioned classifier to bias 
the plan flow toward a requested mode at inference time, we achieve a steer range of \(+0.57\) at the canonical operating point.

\paragraph{The plan is the stronger intervention point.}
A VLA offers two natural alternatives for steering behavior: specifying the
desired decision through the language instruction, or intervening directly on
the generative action expert.

\begin{figure}[t]
  \centering
  \includegraphics[width=0.7\linewidth]{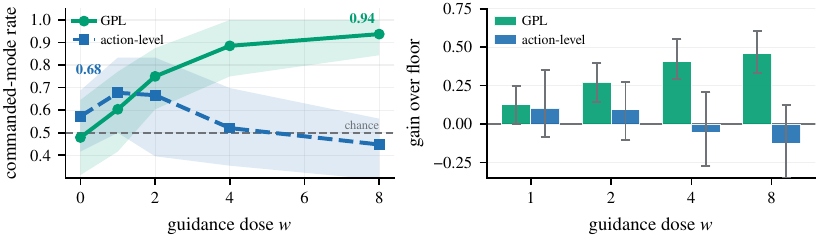}
  \caption{
  \textbf{GPL provides a stronger intervention point than the native action
  distribution.}
   \emph{Left:} commanded-mode rate versus guidance dose for GPL and a
  dedicated no-plan adapter guided through \pizero{}'s native \(128\)-D action
  chunk (\(n=96\) episodes/dose; 95\% cluster bootstrap).
  \emph{Right:} gain over the unguided baseline of each arm, with 95\% intervals.
  }
  \label{fig:action_guidance_native}
\end{figure}

We first compare GPL against guidance of \pizero{}'s native action distribution. Since a generative action expert already represents multiple possible motions, a
central question is whether an additional distribution over plans provides a
meaningfully better control interface. For this action-level baseline, we use the dedicated no-plan adapter, trained on the same task and at the same adaptation budget as the GPL executor.  As shown in \Cref{fig:action_guidance_native}, GPL's commanded-mode rate rises monotonically from 0.48 to 0.94, whereas action-chunk guidance produces no within-arm gain whose interval excludes zero and declines above $w=2$. The relevant decision is explicit in plan space, but is distributed across many action dimensions and timesteps once expressed as a motor trajectory. 

We next evaluate \GPL{} against an ECoT-style language baseline. Both methods issue commands to the identical downstream executor and are evaluated on the same closed-loop success metric: successfully grasping and lifting the specified target cube. Across three adapter seeds and \(40\) held-out spatial configurations, \GPL{} achieves a steering range of \(+0.54\), whereas the language baseline manages only \(+0.02\). We find that downstream execution competence is not the bottleneck: the language baseline actually grasps objects more reliably than the plan-based approach (\(0.50\) versus \(0.33\) overall success rate), yet it systematically fails to follow the specific target command. Furthermore, when opposing steering signals are applied simultaneously, the continuous plan dictates the executed target cube in all \(240\) evaluation rollouts. Equipping the language baseline with a matched trainable projector causes its overall grasp rate to collapse entirely, so its steering range cannot be interpreted (\Cref{app:language_steering}).

\subsection{Outcome Adaptation: Can Experience Change Which Plans Are Selected?}
\label{subsec:si}
While steering specifies a preference externally, outcome adaptation uses
deployment experience to infer which plans should be preferred. If unseen
deployment dynamics invalidate a previously viable strategy, for example, a
larger object causing grasps to fail while pushing remains effective, a policy
that preserves multiple plans does not need to relearn how to act. \GPL{} enables this by reweighting the plan distribution using the agent's own rollout outcomes, shifting toward plans that remain effective.

\begin{figure}[t]
  \centering
  \includegraphics[width=0.85\linewidth]{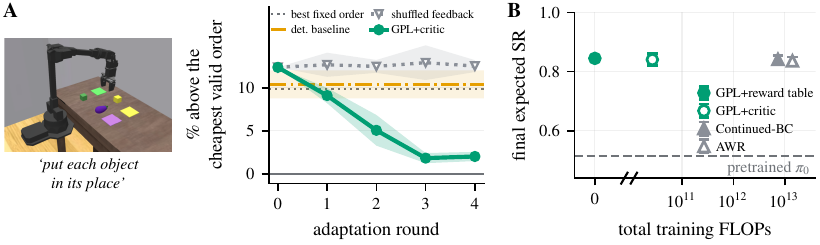}
\caption{
\textbf{Deployment experience improves plan selection. }
\emph{A:} In three-object rearrangement, all orders are valid
but differ in chained end-effector path length; we report normalized regret
above the cheapest order (5 seeds, mean $\pm$ s.d.). The \emph{best fixed
order} is cost-aware but not scene-adaptive.
\emph{B:} In the two-cube task, valid targets have
different unobserved success probabilities; we report final success versus
training FLOPs (10 seeds, 95\% CIs). Frozen-generator reranking matches
retraining baselines with far less compute, while the reward table requires
none.
}
  \label{fig:costsi}
\end{figure}

\paragraph{Experience can optimize among already-valid plans.}
In a three-object rearrangement task, where an object should be placed in its dedicated position identified by the color, the plan dictates the execution order. Because every permutation validly completes the task, the frozen generator learns
them all equally and stays blind to any preference cost. At deployment, we introduce an additional preference that was absent from training: minimizing the chained end-effector path length required to execute the full ordering. We fit a lightweight critic on rollouts from the model itself to score each order. We deploy this as our first task-specific instance of the shared mechanism: a continuous cost regressor to minimize physical penalties. Because the learned \(g_\omega\) is continuous, it can score unseen plans at the cost of fitting \(\omega\). As shown in \Cref{fig:costsi}A, Boltzmann-reranking candidate plans at inference time drastically reduces the expected regret, whereas a shuffled-cost control shows no improvement. Naturally, the deterministic baseline, being a single plan, cannot be reranked.

\paragraph{Why rerank plans rather than action samples?}
Since \pizero{} is a generative policy, a natural alternative is to sample and rerank raw action trajectories directly. A no-plan \pizero{} baseline, matched in scenes, capacity, and critic architecture, fails to reduce execution costs, and the gap favoring plan-level reranking widens as the sample budget ($K$) increases. The reason is executable diversity: action-diffusion noise mostly produces small variations of one committed strategy rather than independently successful alternatives (\Cref{app:rerank_locus}).

\paragraph{Reranking can avoid further generator training.}
Returning to the two-cube task, we introduce a hidden reliability objective: the
two geometrically valid targets have asymmetric, unobserved success
probabilities, and the agent sees only a binary outcome after each grasp. We
rerank frozen plan samples with no generator gradient steps, via either a
\emph{learned critic} (a small regressor fit on the observed outcomes) or a
\emph{reward table} (a smoothed per-mode average, fit-free). As illustrated in
\Cref{fig:costsi}B, both generator-frozen arms match continued behavior
cloning (Continued-BC) and Advantage-Weighted Regression (AWR) \citep{peng2019awr}, both of which
update the flow generator on the same deployment rollouts, within a strict
statistical equivalence margin. The critic reaches parity at
\({\approx}388\times\) fewer training FLOPs, and
the reward table needs none at all.
Construction of the hidden-cost objectives, cross-scene validation, the
reranking-locus ablation, and the FLOP accounting are given in
\Cref{app:group_adaptation}.

\section{When Does the Plan Matter? The Redundancy Principle}
\label{sec:when}

In robotic manipulation, an explicit plan can improve task success only to the extent that it carries
decision-relevant information the executor cannot already read from the
observation. If the observation alone dictates the decision, the plan is redundant. In such cases, its benefits vanish while its interface overhead remains, potentially degrading net success. Conversely, when ambiguity, occlusion, or distributional shifts leave the decision open, the plan becomes load-bearing and must be addressable. We quantify this redundancy with
$\widetilde R = 1-\frac{H(y\mid c)}{\ln m}$,
where \(H(y\mid c)\) is the held-out cross-entropy of a probe predicting the
decision \(y\) from the conditioning \(c\) of the executor, and \(m\) is the number
of modes, assumed marginally uniform (\Cref{app:redundancy-score}). Thus
\(\widetilde R\approx0\) indicates little information about the decision, while
\(\widetilde R\approx1\) indicates it is nearly determined by \(c\).
Three monotone manipulations test whether the effect of the plan tracks this score.

\begin{figure}[t]
  \centering
  \includegraphics[width=0.7\linewidth]{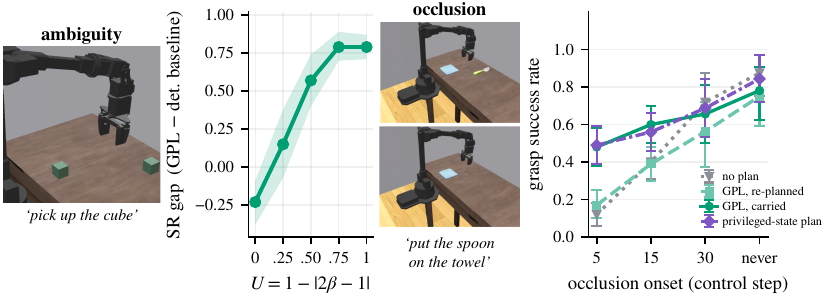}
  \caption{
  \textbf{A plan becomes consequential when it carries unresolved or missing
  decision information.}
  \emph{Left:} the generative advantage grows with controlled ambiguity
  \(U=1-|2\beta-1|\), where \(\beta\) is the fraction of training-demonstration
  mass on the preferred cube.
  \emph{Right:} carrying an earlier decision grows more valuable as occlusion
  removes the information that produced it, relative to stock \pizero{}; the
  privileged-state curve sets a ground-truth-pose ceiling.
  }
  \label{fig:when_plan_matters}
\end{figure}

\paragraph{Ambiguity.}
Sweeping the training distribution under a fixed scene and executor reveals that the generative advantage (success-rate gap over the deterministic baseline) scales directly with task ambiguity (\Cref{fig:when_plan_matters}). Performance shifts from a net penalty in unambiguous tasks to substantial gains under full bimodality. A strictly unimodal control confirms the source of this penalty: when initial observations fully resolve the decision, an intermediate plan provides no new information, acting purely as an architectural bottleneck.

\paragraph{Occlusion and commitment.}
Next, we test temporal redundancy by hiding the target visual mesh from time $t$ onward during a \pizero{} manipulation task. As occlusion destroys decision information in the observation, committing to a previously selected plan becomes critical. Carrying an early plan through the occlusion heavily outperforms both stock \pizero{} and the matched plan-less control (by up to 36 and 28 percentage points, respectively). It also decisively beats re-inferring the plan from the degraded image at every step (\Cref{fig:when_plan_matters}). Importantly, this gain tracks the privileged-state ceiling: when occlusion happens late (or not at all), the observation remains sufficient, the plan becomes redundant again, and the baselines converge.

\paragraph{A causal test on LIBERO.}LIBERO \citep{liu2023libero} provides our sharpest causal test. To strictly isolate whether the \emph{plan} carries the decision, we use generic language instructions that omit specific target names. This requires deploying pretrained \pizerofive{} \citep{black2025pi_05} rather than \pizero{}. Because \pizero{} overfits to verbatim fine-tuning strings, it completely collapses under generic instructions (dropping to just $3/300$ successes), leaving no headroom to measure the effect of the plan. In contrast, \pizerofive{} robustly follows generic commands (retaining $140/300$ successes) owing to its broader pretraining, providing the necessary competent baseline (\Cref{app:redundancy-libero}). However, when the plan generator is trained purely on \pizerofive{}'s default habit rollouts, the resulting plan is largely redundant with the scene ($\widetilde R=0.47$, measured prior to adapter training) and steering is accordingly weak. We intervene by replacing only these habit demonstrations with counterfactual pairs (e.g., placing either bowl in the identical visual context) to deliberately break the observation-decision correlation, holding the executor, benchmark, and adaptation recipe strictly fixed.

Modifying these training demonstrations solely to drive the redundancy score from 0.47 down to 0.00 improves the pooled oracle steer range from a weak effect (+0.19) to a strong effect (+0.65) across all six matched adapter seeds. Furthermore, across a five-point dose sweep constructed from intermediate mixture levels, redundancy and steering performance exhibit a perfect inverse rank correlation (Spearman's $\rho = -1.00$; \Cref{tab:libero-dose}).

Consequently, the unguided learned sampler recovers its steering capability, and counter-habit outcome adaptation becomes positive across all counterfactual seeds (compared to just one of six habit seeds). An ablation that altered instruction wording without changing $\widetilde R$ left the effect unchanged. Ultimately, redundancy predicts whether the executor will \emph{use} the plan, not whether the architecture is globally cost-free: non-redundancy makes the plan valuable, while the conditioning interface itself can impose an independent competence cost (reported in \Cref{sec:limits}).
\section{Generality Across Plan Representations and Domains}
\label{sec:generality}

The addressability results in \Cref{sec:experiments} have so far been
demonstrated only with hand-designed plan coordinates and in a single
simulator, SimplerEnv \cite{li2024evaluating}. We test both dependencies here.

\subsection{Does Addressability Require Hand-Designed Coordinates?}
\label{subsec:learned_coords}

Addressability is not an artifact of hand-designed plan coordinates. While our
previous experiments relied on structured plans with explicitly defined modes,
here we replace those labels with a 2-D plan latent learned entirely from
behavior. We train an autoencoder to compress demonstrated future
end-effector trajectories into this space, using only outer-train layouts and
freezing it before any object, mode, or language labels are introduced. We
then train the same generative flow and deterministic heads over the frozen
learned latents.

\begin{figure}[t]
\centering
\includegraphics[width=0.65\linewidth]{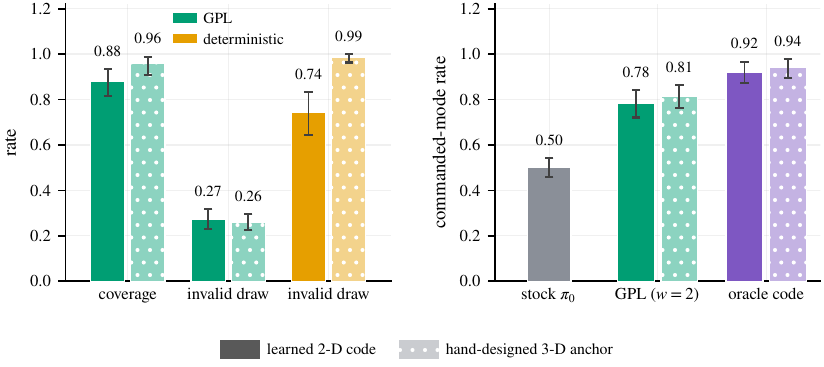}
\caption{\textbf{Addressability survives behavior-learned coordinates.}
\emph{Left:} Offline, properties of samples straight out of the plan
generator, never executed. \emph{Right:} Closed loop (through \pizero{}), the
full manipulation episode is rolled out in the simulator. Error bars are 95\%
percentile-bootstrap intervals over training seeds and held-out layouts
(10,000 replicates).}
\label{fig:learned-coordinates}
\end{figure}

All three addressability operations remain viable without hand-designed
coordinates (\Cref{fig:learned-coordinates}). Offline, the generative flow covers both valid
targets on most held-out scenes, whereas the deterministic readout falls
off-mode on nearly 75\% of its predictions. Thus, despite being learned entirely from data, the resulting latent space remains both
steerable and rerankable. In closed-loop execution through \pizero{}, guiding
the learned plan latent raises the commanded-mode rate from the stock-\pizero{}
baseline of 0.50 to 0.78, approaching the 0.92 oracle-code ceiling obtained
when the executor is given the encoded ground-truth path. This oracle ablation effectively decouples the downstream fidelity of the executor in following the latent representation from the upstream accuracy of the generative flow in producing it.
Most importantly, under identical adapters and scenes, the learned plan latent is
statistically indistinguishable from the hand-designed 3-D anchor in
closed-loop steering (paired difference $+0.03$), despite somewhat lower
offline coverage. Thus, addressability does not require predefined or
human-readable plan coordinates: in this setting, a compact representation
learned directly from demonstrated behavior is sufficient.

\subsection{Schemas, Embodiments, and Domains}
\label{subsec:schemas}

The three capabilities are not tied to one simulator, one embodiment, or one domain. A single 54-D schema of object-relative contact events (position, approach angle, gripper state), trained jointly across four SimplerEnv task families, recovers valid alternatives on all four (\Cref{app:contact-sketch}). On RoboCasa \citep{robocasa2024}, a mobile-manipulator embodiment with a different simulator, scene distribution, and robot, an offline probe with fixed plan anchors shows that \pizero{} realizes both commanded alternatives, and reranking raises plan-level task success from
0.65 to 1.00, though terminal mobile-base navigation remains outside the scope of the adapted
executor (\Cref{app:cross_sim}). Beyond manipulation, the framework extends to an abstract multi-path maze domain using a higher-dimensional route-map plan (\Cref{app:maze}). In this setting, the learned plan space supports coverage, steering, and outcome adaptation over multiple feasible routes through the maze.

\section{Limitations and Scope}
\label{sec:limits}
Three limitations define the current scope of the addressability results. \emph{First}, all experiments are conducted in simulation. The main effects are observed across three simulators and an abstract maze domain, but we have not yet tested them on hardware. \emph{Second}, existing manipulation benchmarks rarely provide the kind of underdetermined decisions needed to study multimodality directly. We therefore construct most of the multimodal settings used here, including the two-cube and three-object tasks, the contact-sketch scenes, and the LIBERO counterfactual demonstrations. Pretrained VLAs do not always adapt well to this shift. On LIBERO, \pizero{} achieves only $3/300$ successes under the generic instructions required by the test, which motivates our switch to \pizerofive{}. Even with \pizerofive{}, none of the four preregistered magnitude predictions are met. Removing the redundancy hypothesized to drive this degradation also does not remove the full performance cost: the plan-conditioning interface retains a baseline grasp-rate penalty of $0.274$. \emph{Third}, we validated the redundancy principle across three causal manipulations: an
ambiguity sweep, an occlusion sweep, and the LIBERO counterfactual test. We do not yet know whether the same principle extends to other robot task families or to reasoning VLAs more broadly.
\section{Conclusion}
\label{sec:conclusion}

We investigate where multimodality should live in a reasoning-augmented VLA.
A generative action model can represent multiple motor trajectories without preserving distinct task-level choices. GPL instead represents the intermediate decision as a distribution over
plans, keeping those alternatives available before execution. This makes the
reasoning state \emph{addressable}: valid decisions can be sampled, a requested
one can be selected, and deployment outcomes can change which alternatives are
preferred without retraining the underlying VLA.

Across our experiments, the plan distribution provides a stronger control interface than language prompting or intervention on \pizero{}'s native action distribution. The same representation also supports adaptation from experience. Finally, the redundancy principle clarifies when this interface should matter: an explicit plan becomes consequential when it carries decision information that the observation does not already determine. Together, these results suggest that intermediate reasoning in VLAs should be designed not only to encode a decision; it must establish \emph{addressability}, exposing the full space of viable alternatives over which that decision can still be made.

\subsection*{AI use statement}

In this work, we used generative AI tools for creating and editing software code, drafting parts of the research paper, editing the manuscript to improve readability, summarizing and analyzing existing literature, and the creation of the code for figures.

We have reviewed all AI-assisted work. LLM-generated summaries of existing literature were cross-referenced with the original papers to ensure factual accuracy and screen for potential hallucination. All AI-assisted software code, including implementations of model architectures, training loops, and data pipelines, was tested for correctness and manually verified by the authors. Edits made to improve the readability of the text were reviewed line by line to check that they preserved the original scientific intent. We take responsibility for the final content of this work, including text, claims, or artifacts produced with the aid of generative AI.

\subsection*{Reproducibility statement}

Every headline number in this paper states the variance unit it was computed over (held-out scenes, layouts, or episodes for the confirmatory family; training or rollout seeds elsewhere) at the point it is reported. The appendix gives a complete reproducibility record for every experiment: \Cref{app:gpl_training,app:pi0} the GPL training and \pizero{}/\pizerofive{} adaptation recipes; \Cref{app:lara,app:learned_coords} the matched-baseline and learned-coordinate recipes; \Cref{app:maze,app:maze_impl} the maze dataset generators and implementation; \Cref{app:group_steering,app:group_adaptation} the full protocol, baselines, and controls behind each addressability capability, including the \Cref{app:frontier}{} compute-frontier FLOP accounting; \Cref{app:redundancy-libero} the redundancy-score estimators and LIBERO counterfactual intervention; and \Cref{app:contact-sketch,app:cross_sim} the multi-family and cross-simulator generality battery. Every recipe table reports optimizer, learning rate, batch size, and seeds. Code will be released at camera-ready.

\bibliography{references}

\begin{thebibliography}{48}
\providecommand{\natexlab}[1]{#1}
\providecommand{\url}[1]{\texttt{#1}}
\expandafter\ifx\csname urlstyle\endcsname\relax
  \providecommand{\doi}[1]{doi: #1}\else
  \providecommand{\doi}{doi: \begingroup \urlstyle{rm}\Url}\fi

\bibitem[Ajay et~al.(2023)Ajay, Du, Gupta, Tenenbaum, Jaakkola, and
  Agrawal]{ajay2023is}
Anurag Ajay, Yilun Du, Abhi Gupta, Joshua~B. Tenenbaum, Tommi~S. Jaakkola, and
  Pulkit Agrawal.
\newblock Is conditional generative modeling all you need for decision making?
\newblock In \emph{The Eleventh International Conference on Learning
  Representations}, 2023.
\newblock URL \url{https://openreview.net/forum?id=sP1fo2K9DFG}.

\bibitem[Bai et~al.(2026)Bai, Lyu, Zhou, Li, Wang, Xing, Zhao, Wang, Wang, Chi,
  et~al.]{bai2026latent}
Shuanghao Bai, Jing Lyu, Wanqi Zhou, Zhe Li, Dakai Wang, Lei Xing, Xiaoguang
  Zhao, Pengwei Wang, Zhongyuan Wang, Cheng Chi, et~al.
\newblock Latent reasoning vla: Latent thinking and prediction for
  vision-language-action models.
\newblock \emph{arXiv preprint arXiv:2602.01166}, 2026.

\bibitem[Beyer et~al.(2024)Beyer, Steiner, Pinto, Kolesnikov, Wang, Salz,
  Neumann, Alabdulmohsin, Tschannen, Bugliarello, et~al.]{beyer2024paligemma}
Lucas Beyer, Andreas Steiner, Andr{\'e}~Susano Pinto, Alexander Kolesnikov,
  Xiao Wang, Daniel Salz, Maxim Neumann, Ibrahim Alabdulmohsin, Michael
  Tschannen, Emanuele Bugliarello, et~al.
\newblock Paligemma: A versatile 3b vlm for transfer.
\newblock \emph{arXiv preprint arXiv:2407.07726}, 2024.

\bibitem[Black et~al.(2024{\natexlab{a}})Black, Brown, Driess, Esmail, Equi,
  Finn, Fusai, Groom, Hausman, Ichter, et~al.]{black2024pi_0}
Kevin Black, Noah Brown, Danny Driess, Adnan Esmail, Michael Equi, Chelsea
  Finn, Niccolo Fusai, Lachy Groom, Karol Hausman, Brian Ichter, et~al.
\newblock $\pi_0$: A vision-language-action flow model for general robot
  control.
\newblock \emph{arXiv preprint arXiv:2410.24164}, 2024{\natexlab{a}}.

\bibitem[Black et~al.(2024{\natexlab{b}})Black, Nakamoto, Atreya, Walke, Finn,
  Kumar, and Levine]{black2024zero}
Kevin Black, Mitsuhiko Nakamoto, Pranav Atreya, Homer Walke, Chelsea Finn,
  Aviral Kumar, and Sergey Levine.
\newblock Zero-shot robotic manipulation with pretrained image-editing
  diffusion models.
\newblock In \emph{International Conference on Learning Representations
  (ICLR)}, 2024{\natexlab{b}}.

\bibitem[Black et~al.(2025)Black, Brown, Darpinian, Dhabalia, Driess, Esmail,
  Equi, Finn, Fusai, Galliker, et~al.]{black2025pi_05}
Kevin Black, Noah Brown, James Darpinian, Karan Dhabalia, Danny Driess, Adnan
  Esmail, Michael Equi, Chelsea Finn, Niccolo Fusai, Manuel~Y. Galliker, et~al.
\newblock $\pi_{0.5}$: a vision-language-action model with open-world
  generalization.
\newblock \emph{arXiv preprint arXiv:2504.16054}, 2025.

\bibitem[Chen et~al.(2025)Chen, Tian, Liu, Zhou, Li, and Zhao]{chen2025conrft}
Yuhui Chen, Shuai Tian, Shugao Liu, Yingting Zhou, Haoran Li, and Dongbin Zhao.
\newblock Conrft: A reinforced fine-tuning method for vla models via
  consistency policy.
\newblock \emph{arXiv preprint arXiv:2502.05450}, 2025.

\bibitem[Chi et~al.(2025)Chi, Xu, Feng, Cousineau, Du, Burchfiel, Tedrake, and
  Song]{chi2025diffusion}
Cheng Chi, Zhenjia Xu, Siyuan Feng, Eric Cousineau, Yilun Du, Benjamin
  Burchfiel, Russ Tedrake, and Shuran Song.
\newblock Diffusion policy: Visuomotor policy learning via action diffusion.
\newblock \emph{The International Journal of Robotics Research}, 44\penalty0
  (10-11):\penalty0 1684--1704, 2025.

\bibitem[Du et~al.(2023)Du, Yang, Dai, Dai, Nachum, Tenenbaum, Schuurmans, and
  Abbeel]{du2023learning}
Yilun Du, Sherry Yang, Bo~Dai, Hanjun Dai, Ofir Nachum, Joshua~B. Tenenbaum,
  Dale Schuurmans, and Pieter Abbeel.
\newblock Learning universal policies via text-guided video generation.
\newblock In \emph{Advances in Neural Information Processing Systems
  (NeurIPS)}, 2023.

\bibitem[Florence et~al.(2022)Florence, Lynch, Zeng, Ramirez, Wahid, Downs,
  Wong, Lee, Mordatch, and Tompson]{florence2022implicit}
Pete Florence, Corey Lynch, Andy Zeng, Oscar~A Ramirez, Ayzaan Wahid, Laura
  Downs, Adrian Wong, Johnny Lee, Igor Mordatch, and Jonathan Tompson.
\newblock Implicit behavioral cloning.
\newblock In \emph{Conference on robot learning}, pp.\  158--168. PMLR, 2022.

\bibitem[Guzm{\'a}n-Rivera et~al.(2012)Guzm{\'a}n-Rivera, Batra, and
  Kohli]{guzmanrivera2012multiple}
Abner Guzm{\'a}n-Rivera, Dhruv Batra, and Pushmeet Kohli.
\newblock Multiple choice learning: Learning to produce multiple structured
  outputs.
\newblock In \emph{Advances in Neural Information Processing Systems
  (NeurIPS)}, 2012.

\bibitem[Hao et~al.(2024)Hao, Sukhbaatar, Su, Li, Hu, Weston, and
  Tian]{hao2024training}
Shibo Hao, Sainbayar Sukhbaatar, DiJia Su, Xian Li, Zhiting Hu, Jason Weston,
  and Yuandong Tian.
\newblock Training large language models to reason in a continuous latent
  space.
\newblock \emph{arXiv preprint arXiv:2412.06769}, 2024.

\bibitem[Ho \& Salimans(2022)Ho and
  Salimans]{ho2022classifierfreediffusionguidance}
Jonathan Ho and Tim Salimans.
\newblock Classifier-free diffusion guidance, 2022.
\newblock URL \url{https://arxiv.org/abs/2207.12598}.

\bibitem[Hu et~al.(2021)Hu, Shen, Wallis, Allen-Zhu, Li, Wang, Wang, and
  Chen]{hu2021lora}
Edward~J Hu, Yelong Shen, Phillip Wallis, Zeyuan Allen-Zhu, Yuanzhi Li, Shean
  Wang, Lu~Wang, and Weizhu Chen.
\newblock {Lora}: Low-rank adaptation of large language models.
\newblock \emph{arXiv preprint arXiv:2106.09685}, 2021.

\bibitem[Huang et~al.(2026{\natexlab{a}})Huang, Man, Yu, Chen, Kautz, Wang, and
  Yang]{huang2026fast}
Chi-Pin Huang, Yunze Man, Zhiding Yu, Min-Hung Chen, Jan Kautz, Yu-Chiang~Frank
  Wang, and Fu-En Yang.
\newblock Fast-thinkact: Efficient vision-language-action reasoning via
  verbalizable latent planning.
\newblock \emph{arXiv preprint arXiv:2601.09708}, 2026{\natexlab{a}}.

\bibitem[Huang et~al.(2026{\natexlab{b}})Huang, Wu, Chen, Wang, and
  Yang]{huang2026thinkact}
Chi-Pin Huang, Yueh-Hua Wu, Min-Hung Chen, Frank Wang, and Fu-En Yang.
\newblock Thinkact: Vision-language-action reasoning via reinforced visual
  latent planning.
\newblock \emph{Advances in Neural Information Processing Systems},
  38:\penalty0 82782--82802, 2026{\natexlab{b}}.

\bibitem[Huang et~al.(2025)Huang, Chen, Wang, Li, and Qi]{huang2025reinforcing}
Zemin Huang, Zhiyang Chen, Zijun Wang, Tiancheng Li, and Guo-Jun Qi.
\newblock Reinforcing the diffusion chain of lateral thought with diffusion
  language models.
\newblock \emph{arXiv preprint arXiv:2505.10446}, 2025.

\bibitem[Janner et~al.(2022)Janner, Du, Tenenbaum, and
  Levine]{janner2022planning}
Michael Janner, Yilun Du, Joshua~B. Tenenbaum, and Sergey Levine.
\newblock Planning with diffusion for flexible behavior synthesis.
\newblock In \emph{International Conference on Machine Learning (ICML)}, 2022.

\bibitem[Kang et~al.(2026)Kang, Zhang, Kuang, Majamaki, Jaitly, Ma, and
  Qin]{kang2026ladirlatentdiffusionenhances}
Haoqiang Kang, Yizhe Zhang, Nikki~Lijing Kuang, Nicklas Majamaki, Navdeep
  Jaitly, Yi-An Ma, and Lianhui Qin.
\newblock Ladir: Latent diffusion enhances llms for text reasoning, 2026.
\newblock URL \url{https://arxiv.org/abs/2510.04573}.

\bibitem[Kwok et~al.(2025)Kwok, Agia, Sinha, Foutter, Li, Stoica, Mirhoseini,
  and Pavone]{kwok2025robomonkeyscalingtesttimesampling}
Jacky Kwok, Christopher Agia, Rohan Sinha, Matt Foutter, Shulu Li, Ion Stoica,
  Azalia Mirhoseini, and Marco Pavone.
\newblock Robomonkey: Scaling test-time sampling and verification for
  vision-language-action models, 2025.
\newblock URL \url{https://arxiv.org/abs/2506.17811}.

\bibitem[Lee et~al.(2024)Lee, Wang, Etukuru, Kim, Shafiullah, and
  Pinto]{lee2024behavior}
Seungjae Lee, Yibin Wang, Haritheja Etukuru, H~Jin Kim, Nur Muhammad~Mahi
  Shafiullah, and Lerrel Pinto.
\newblock Behavior generation with latent actions.
\newblock \emph{arXiv preprint arXiv:2403.03181}, 2024.

\bibitem[Li et~al.(2025)Li, Zuo, Yu, Zhang, Yang, Zhang, Zhu, Zhang, Chen, Cui,
  et~al.]{li2025simplevla}
Haozhan Li, Yuxin Zuo, Jiale Yu, Yuhao Zhang, Zhaohui Yang, Kaiyan Zhang,
  Xuekai Zhu, Yuchen Zhang, Tianxing Chen, Ganqu Cui, et~al.
\newblock Simplevla-rl: Scaling vla training via reinforcement learning.
\newblock \emph{arXiv preprint arXiv:2509.09674}, 2025.

\bibitem[Li et~al.(2024)Li, Hsu, Gu, Mees, Pertsch, Walke, Fu, Lunawat, Sieh,
  Kirmani, Levine, Wu, Finn, Su, Vuong, and Xiao]{li2024evaluating}
Xuanlin Li, Kyle Hsu, Jiayuan Gu, Oier Mees, Karl Pertsch, Homer~Rich Walke,
  Chuyuan Fu, Ishikaa Lunawat, Isabel Sieh, Sean Kirmani, Sergey Levine, Jiajun
  Wu, Chelsea Finn, Hao Su, Quan Vuong, and Ted Xiao.
\newblock Evaluating real-world robot manipulation policies in simulation.
\newblock In \emph{8th Annual Conference on Robot Learning}, 2024.
\newblock URL \url{https://openreview.net/forum?id=LZh48DTg71}.

\bibitem[Lipman et~al.(2023)Lipman, Chen, Ben-Hamu, Nickel, and
  Le]{lipman2023flow}
Yaron Lipman, Ricky T.~Q. Chen, Heli Ben-Hamu, Maximilian Nickel, and Matthew
  Le.
\newblock Flow matching for generative modeling.
\newblock In \emph{The Eleventh International Conference on Learning
  Representations}, 2023.
\newblock URL \url{https://openreview.net/forum?id=PqvMRDCJT9t}.

\bibitem[Liu et~al.(2023{\natexlab{a}})Liu, Zhu, Gao, Feng, qiang liu, Zhu, and
  Stone]{liu2023libero}
Bo~Liu, Yifeng Zhu, Chongkai Gao, Yihao Feng, qiang liu, Yuke Zhu, and Peter
  Stone.
\newblock {LIBERO}: Benchmarking knowledge transfer for lifelong robot
  learning.
\newblock In \emph{Thirty-seventh Conference on Neural Information Processing
  Systems Datasets and Benchmarks Track}, 2023{\natexlab{a}}.
\newblock URL \url{https://openreview.net/forum?id=xzEtNSuDJk}.

\bibitem[Liu et~al.(2023{\natexlab{b}})Liu, Gong, and qiang liu]{liu2023flow}
Xingchao Liu, Chengyue Gong, and qiang liu.
\newblock Flow straight and fast: Learning to generate and transfer data with
  rectified flow.
\newblock In \emph{The Eleventh International Conference on Learning
  Representations}, 2023{\natexlab{b}}.
\newblock URL \url{https://openreview.net/forum?id=XVjTT1nw5z}.

\bibitem[Lovelace et~al.(2025)Lovelace, Belardi, Zalouk, Polavaram, Kundurthy,
  and Weinberger]{lovelace2025stopthinkautoregress}
Justin Lovelace, Christian~K Belardi, Sofian Zalouk, Adhitya Polavaram,
  Srivatsa~R Kundurthy, and Kilian~Q Weinberger.
\newblock Stop-think-autoregress: Language modeling with latent diffusion
  planning.
\newblock In \emph{Second Conference on Language Modeling}, 2025.
\newblock URL \url{https://openreview.net/forum?id=c05qIG1Z2B}.

\bibitem[Lu et~al.(2026)Lu, Qin, Jiang, Wang, Zhang, Liang, Yang, Xu, and
  Zhao]{lu2026unisteerunifiednoisesteering}
Junjie Lu, Xinyao Qin, Yuhua Jiang, Kaixin Wang, Chuheng Zhang, Bin Liang, Jun
  Yang, Min Xu, and Li~Zhao.
\newblock Unisteer: Unified noise steering for efficient human-guided vla
  adaptation, 2026.
\newblock URL \url{https://arxiv.org/abs/2605.10821}.

\bibitem[Lynch \& Sermanet(2021)Lynch and Sermanet]{lynch2021language}
Corey Lynch and Pierre Sermanet.
\newblock Language conditioned imitation learning over unstructured data.
\newblock In \emph{Robotics: Science and Systems (RSS)}, 2021.

\bibitem[Lynch et~al.(2019)Lynch, Khansari, Xiao, Kumar, Tompson, Levine, and
  Sermanet]{lynch2019learning}
Corey Lynch, Mohi Khansari, Ted Xiao, Vikash Kumar, Jonathan Tompson, Sergey
  Levine, and Pierre Sermanet.
\newblock Learning latent plans from play.
\newblock In \emph{Conference on Robot Learning (CoRL)}, 2019.

\bibitem[Mees et~al.(2022)Mees, Hermann, and Burgard]{mees2022hulc}
Oier Mees, Lukas Hermann, and Wolfram Burgard.
\newblock What matters in language conditioned robotic imitation learning over
  unstructured data.
\newblock \emph{IEEE Robotics and Automation Letters (RA-L)}, 7\penalty0
  (4):\penalty0 11205--11212, 2022.

\bibitem[Nasiriany et~al.(2024)Nasiriany, Maddukuri, Zhang, Parikh, Lo, Joshi,
  Mandlekar, and Zhu]{robocasa2024}
Soroush Nasiriany, Abhiram Maddukuri, Lance Zhang, Adeet Parikh, Aaron Lo,
  Abhishek Joshi, Ajay Mandlekar, and Yuke Zhu.
\newblock Robocasa: Large-scale simulation of everyday tasks for generalist
  robots.
\newblock In \emph{Robotics: Science and Systems (RSS)}, 2024.

\bibitem[Peng et~al.(2019)Peng, Kumar, Zhang, and Levine]{peng2019awr}
Xue~Bin Peng, Aviral Kumar, Grace Zhang, and Sergey Levine.
\newblock Advantage-weighted regression: Simple and scalable off-policy
  reinforcement learning, 2019.
\newblock URL \url{https://arxiv.org/abs/1910.00177}.

\bibitem[Rosete-Beas et~al.(2022)Rosete-Beas, Mees, Kalweit, Boedecker, and
  Burgard]{rosete2022tacorl}
Erick Rosete-Beas, Oier Mees, Gabriel Kalweit, Joschka Boedecker, and Wolfram
  Burgard.
\newblock Latent plans for task agnostic offline reinforcement learning.
\newblock In \emph{Proceedings of the 6th Conference on Robot Learning (CoRL)},
  2022.

\bibitem[Rupprecht et~al.(2017)Rupprecht, Laina, DiPietro, Baust, Tombari,
  Navab, and Hager]{Rupprecht_2017_ICCV}
Christian Rupprecht, Iro Laina, Robert DiPietro, Maximilian Baust, Federico
  Tombari, Nassir Navab, and Gregory~D. Hager.
\newblock Learning in an uncertain world: Representing ambiguity through
  multiple hypotheses.
\newblock In \emph{Proceedings of the IEEE International Conference on Computer
  Vision (ICCV)}, Oct 2017.

\bibitem[Schwarzschild et~al.(2021)Schwarzschild, Borgnia, Gupta, Huang,
  Vishkin, Goldblum, and Goldstein]{schwarzschild2021can}
Avi Schwarzschild, Eitan Borgnia, Arjun Gupta, Furong Huang, Uzi Vishkin, Micah
  Goldblum, and Tom Goldstein.
\newblock Can you learn an algorithm? generalizing from easy to hard problems
  with recurrent networks.
\newblock \emph{Advances in Neural Information Processing Systems},
  34:\penalty0 6695--6706, 2021.

\bibitem[Song et~al.(2020)Song, Meng, and Ermon]{song2020denoising}
Jiaming Song, Chenlin Meng, and Stefano Ermon.
\newblock Denoising diffusion implicit models.
\newblock \emph{arXiv:2010.02502}, October 2020.
\newblock URL \url{https://arxiv.org/abs/2010.02502}.

\bibitem[Song et~al.(2026)Song, Le, Park, Wang, Shi, Liu, Gu, Eaton, Jayaraman,
  and Daniilidis]{song2026omniguide}
Yunzhou Song, Long Le, Yong-Hyun Park, Jie Wang, Junyao Shi, Lingjie Liu,
  Jiatao Gu, Eric Eaton, Dinesh Jayaraman, and Kostas Daniilidis.
\newblock Omniguide: Universal guidance fields for enhancing generalist robot
  policies.
\newblock \emph{arXiv preprint arXiv:2603.10052}, 2026.

\bibitem[Tan et~al.(2025)Tan, Dou, Zhao, and Krähenbühl]{tan2025interactive}
Shuhan Tan, Kairan Dou, Yue Zhao, and Philipp Krähenbühl.
\newblock Interactive post-training for vision-language-action models, 2025.
\newblock URL \url{https://arxiv.org/abs/2505.17016}.

\bibitem[Tang et~al.(2026)Tang, Chen, Wagenmaker, Finn, and
  Levine]{tang2026improving}
Andy Tang, William Chen, Andrew Wagenmaker, Chelsea Finn, and Sergey Levine.
\newblock Improving robotic generalist policies via flow reversal steering.
\newblock \emph{arXiv preprint arXiv:2606.13675}, 2026.

\bibitem[Tu et~al.(2026)Tu, Fu, Yu, Tang, Kang, Qin, Zhang, and
  Gu]{tu2026latent}
Guancheng Tu, Xiangjun Fu, Suhao Yu, Yao Tang, Haoqiang Kang, Lianhui Qin,
  Yizhe Zhang, and Jiatao Gu.
\newblock Latent reasoning with normalizing flows.
\newblock \emph{arXiv preprint arXiv:2606.06447}, 2026.

\bibitem[Tur et~al.(2026)Tur, Naghiyev, Fang, Tsai, Duan, Fox, and
  Krishna]{tur2026rdvla}
Yalcin Tur, Jalal Naghiyev, Haoquan Fang, Wei-Chuan Tsai, Jiafei Duan, Dieter
  Fox, and Ranjay Krishna.
\newblock Recurrent-depth vla: Implicit test-time compute scaling of
  vision-language-action models via latent iterative reasoning, 2026.
\newblock URL \url{https://arxiv.org/abs/2602.07845}.

\bibitem[Venkatraman et~al.(2024)Venkatraman, Khaitan, Akella, Dolan,
  Schneider, and Berseth]{venkatraman2024reasoning}
Siddarth Venkatraman, Shivesh Khaitan, Ravi~Tej Akella, John Dolan, Jeff
  Schneider, and Glen Berseth.
\newblock Reasoning with latent diffusion in offline reinforcement learning.
\newblock In \emph{The Twelfth International Conference on Learning
  Representations}, 2024.
\newblock URL \url{https://openreview.net/forum?id=tGQirjzddO}.

\bibitem[Wagenmaker et~al.(2025)Wagenmaker, Nakamoto, Zhang, Park, Yagoub,
  Nagabandi, Gupta, and Levine]{wagenmaker2025steering}
Andrew Wagenmaker, Mitsuhiko Nakamoto, Yunchu Zhang, Seohong Park, Waleed
  Yagoub, Anusha Nagabandi, Abhishek Gupta, and Sergey Levine.
\newblock Steering your diffusion policy with latent space reinforcement
  learning.
\newblock \emph{Conference on Robot Learning}, 2025.

\bibitem[Ye et~al.(2024)Ye, Gong, Chen, Zheng, Gao, Shi, Wu, Jiang, Li, Bi, and
  Kong]{ye2024diffusion}
Jiacheng Ye, Shansan Gong, Liheng Chen, Lin Zheng, Jiahui Gao, Han Shi, Chuan
  Wu, Xin Jiang, Zhenguo Li, Wei Bi, and Lingpeng Kong.
\newblock Diffusion of thought: Chain-of-thought reasoning in diffusion
  language models.
\newblock \emph{Advances in Neural Information Processing Systems},
  37:\penalty0 105345--105374, 2024.

\bibitem[Zawalski et~al.(2024)Zawalski, Chen, Pertsch, Mees, Finn, and
  Levine]{Zawalski24-ecot}
Michał Zawalski, William Chen, Karl Pertsch, Oier Mees, Chelsea Finn, and
  Sergey Levine.
\newblock Robotic control via embodied chain-of-thought reasoning.
\newblock \emph{arXiv preprint arXiv:2407.08693}, 2024.

\bibitem[Zhao et~al.(2025)Zhao, Lu, Kim, Fu, Zhang, Wu, Li, Ma, Han, Finn,
  et~al.]{zhao2025cot}
Qingqing Zhao, Yao Lu, Moo~Jin Kim, Zipeng Fu, Zhuoyang Zhang, Yecheng Wu,
  Zhaoshuo Li, Qianli Ma, Song Han, Chelsea Finn, et~al.
\newblock Cot-vla: Visual chain-of-thought reasoning for vision-language-action
  models.
\newblock \emph{arXiv preprint arXiv:2503.22020}, 2025.

\bibitem[Zheng et~al.(2026)Zheng, Marri, Choudhuri, Planche, Gao, Nguyen, Chen,
  Chowdhary, and Wu]{zheng2026failing}
Meng Zheng, Samhita Marri, Anwesa Choudhuri, Benjamin Planche, Zhongpai Gao,
  Van~Nguyen Nguyen, Terrence Chen, Girish Chowdhary, and Ziyan Wu.
\newblock Failing forward: Adaptive failure-informed learning for
  vision-language-action models, 2026.
\newblock URL \url{https://arxiv.org/abs/2605.08434}.

\end{thebibliography}
\bibliographystyle{iclr2027_conference}

\appendix
\newpage 

\section{Theory and Related-Work Comparison}
\subsection{A Committed Regression Readout Mode-Averages}
\label{app:theory}

\begin{proposition}[A committed regression readout mode-averages]
\label{prop:mean}
Let $c$ denote the conditioning information and let $z\in\mathbb{R}^{d_z}$ denote
a plan drawn from $p(z\mid c)$. A deterministic readout $\hat z(c)$ trained with
squared error, $\min_{\hat z}\mathbb{E}_{c,z}\|\hat z(c)-z\|_2^2$, has
population minimizer $\hat z^{*}(c)=\mathbb{E}[z\mid c]\equiv\zdet$.
Suppose that, for a fixed $c$, the valid plans are the distinct modes
$\{z_1,\ldots,z_m\}$, with $m\geq 2$ and probabilities $\pi_i>0$. Then
\begin{equation}
    \zdet
    =
    \sum_{i=1}^{m}\pi_i z_i
    \in
    \operatorname{conv}\{z_1,\ldots,z_m\},
\end{equation}
and if the modes are affinely independent (in particular, whenever $m=2$), this
strict convex combination is not itself a valid mode:
$\zdet \notin \{z_1,\ldots,z_m\}$ and
$\Delta(c)\equiv\min_i \left\|\zdet-z_i\right\|_2>0$.
\end{proposition}

\begin{corollary}[What a committed point cannot expose]
\label{cor:point}
For a fixed observation, a deterministic plan readout exposes at most one of
$m$ valid modes, with zero probability mass on any other; by
\Cref{prop:mean}, that one point can itself be the invalid mean. There is
therefore nothing for plan-level reweighting (steering) or outcome adaptation
(reranking) to act on.
\end{corollary}

\subsection{Where Latent Reasoning Appears in Action-Domain VLAs}
\label{app:related_table}

 \Cref{tab:cell} compares how existing action-domain VLAs represent intermediate
  decisions and whether those decisions remain addressable at inference.

\begin{table}[h]
\centering
\caption{
\textbf{Where latent reasoning appears in action-domain VLAs, and which of them is addressable.}
 The comparison concerns the intermediate decision supplied to the action model,
  not whether the downstream action head is generative. Among the listed methods, \GPL{} is the only method that is addressable.
}
\label{tab:cell}
\small
\setlength{\tabcolsep}{5pt}
\begin{tabular}{lcccc}
\toprule
method & continuous latent & iterative/nonlinear & generative plan & addressable \\
\midrule
LaRA-VLA \citep{bai2026latent}
    & \checkmark & -- & -- & -- \\
ThinkAct \citep{huang2026thinkact}
    & \checkmark & -- & -- & -- \\
Fast-ThinkAct \citep{huang2026fast}
    & \checkmark & -- & -- & -- \\
Recurrent-Depth \citep{tur2026rdvla}
    & \checkmark & \checkmark & -- & -- \\
CoT-VLA \citep{zhao2025cot}
    & -- (pixels) & \checkmark & \checkmark & -- \\
\textbf{\GPL{} (Ours)}
    & \checkmark & \checkmark & \checkmark & \checkmark \\
\bottomrule
\end{tabular}
\end{table}

\section{Setup and Reproducibility}
Gives the full reproducibility record referenced throughout the main text: the GPL training and sampling recipe (\Cref{app:gpl_training}), how a pretrained VLA is adapted to receive a plan (\Cref{app:pi0}), the matched LaRA-style baseline (\Cref{app:lara}), and the learned-coordinate control (\Cref{app:learned_coords}).

\subsection{GPL Training and Sampling}
\label{app:gpl_training}
GPL models a distribution over task-level plans rather than directly over actions. Training is performed entirely offline from pairs $(c,z)$, where $c$ is a frozen representation of the current observation and instruction and $z$ is a low-dimensional plan extracted from the corresponding demonstration.

\paragraph{Conditioning representation.}
For each observation--instruction pair, we run the frozen reasoning backbone once with hidden-state outputs enabled and cache the last transformer's final-token representation,
\[c=f_{\phi}(o,l)\in\mathbb{R}^{d_{\text{model}}}.\]
These features are precomputed for the complete training set and stored offline. GPL training therefore does not require backpropagation through the pretrained VLA.

\paragraph{Rectified-flow generator.}
Let $z\in\mathbb{R}^{d_z}$ denote the plan extracted from a demonstration. We train a conditional velocity field $v_{\theta}(z_t,t,c)$ using rectified flow. For
\[z_0\sim\mathcal{N}(0,I),\qquad t\sim\mathcal{U}[0,1],\qquad z_t=(1-t)z_0+t z,\]
the training objective is
\[\mathcal{L}_{\mathrm{GPL}} = \mathbb{E} \left[ \left\Vert{} v_{\theta}(z_t,t,c)-(z-z_0) \right\Vert{}_2^2 \right].\]
The conditioning vector is first projected from $d_{\text{model}}$ to 64 dimensions using a linear layer, LayerNorm, and GELU. Time is represented with a 64-dimensional sinusoidal embedding. The noisy plan, time embedding, and projected conditioning vector are concatenated and passed through a two-hidden-layer MLP with width 256 and SiLU activations:
\[[d_z+64+64] \rightarrow 256 \rightarrow 256 \rightarrow d_z.\]
\Cref{tab:gpl_recipe} summarizes the full recipe.

\begin{table}[ht]
\centering
\small
\caption{\textbf{GPL training recipe.} The frozen backbone is used only to precompute conditioning features; the generative model itself is a small MLP trained from scratch for each plan representation.}
\label{tab:gpl_recipe}
\begin{tabular}{ll}
\toprule
\textbf{Component} & \textbf{Configuration} \\
\midrule
Conditioning $c$ & Frozen VLA final-layer, final-token feature \\
Conditioning dimension & $d_{\text{model}}$ \\
Condition encoder & Linear $d_{\text{model}}\!\rightarrow\!64$, LayerNorm, GELU \\
Time embedding & 64-D sinusoidal \\
Flow hidden layers & 256, 256 \\
Flow activation & SiLU \\
Noise distribution & $\mathcal{N}(0,I)$ \\
Time distribution & $\mathcal{U}[0,1]$ \\
Objective & Rectified-flow velocity matching \\
Optimizer & Adam \\
Learning rate & $2\times10^{-3}$ \\
Batch size & 128 \\
Optimization steps & 4000 \\
Sampling solver & Euler \\
Sampling steps & {4,6,8} \\
\bottomrule
\end{tabular}
\end{table}

\paragraph{Sampling.}
At inference time, we initialize $z_0\sim\mathcal{N}(0,I)$ and numerically integrate the learned velocity field from $t=0$ to $t=1$ using Euler updates,
\[z_{i+1} = z_i+\Delta t\, v_{\theta}(z_i,t_i,c).\]
Unless otherwise noted, we use a fixed-step Euler solver with $n=4$ steps for GPL sampling. When applying classifier guidance, we increase the number of integration steps to $n \in \{6, 8\}$ to allow the guidance gradient sufficient iterations to effectively steer the generation.

\subsection{VLA Adaptation and Plan Injection}
\label{app:pi0}
We condition the pretrained VLA on a plan latent $z$ through learned soft tokens while keeping the pretrained backbone frozen. The dimensionality of $z$ is task-dependent; examples used in our experiments include object position, grasp orientation, and higher-dimensional contact sketches. The conditioning mechanism itself is independent of this dimensionality.

\paragraph{Plan projection and injection.}
A small MLP $P_{\eta}$ maps the plan latent to $k=8$ embeddings with the same width as the VLA prefix:
\[
P_{\eta}\colon \mathbb{R}^{d_z} \rightarrow \mathbb{R}^{8\times 2048}.
\]
Concretely, the projector consists of
\[
d_z \rightarrow 512 \rightarrow 512 \rightarrow 8\!\times\!2048,
\]
with SiLU nonlinearities between linear layers. The resulting embeddings are appended to the image--language prefix constructed by the pretrained PaliGemma backbone \citep{beyer2024paligemma}. They participate in the same bidirectional prefix attention as the original image and language tokens. The same modified prefix construction is used during both training and action sampling, so no separate inference-time conditioning mechanism is required.

All parameters of the pretrained VLA are frozen. We optimize the plan projector together with a lightweight LoRA adaptation of the action expert. Training uses the VLA's original flow-matching objective, with no additional plan-specific loss. The precise checkpoint, LoRA parameterization, and optimization hyperparameters are domain-specific and are reported in Table~\ref{tab:vla_adaptation_recipe}. With probability \(p_{\mathrm{drop}}\), the plan is omitted during training, providing a classifier-free anchor to the plan-less policy (Classifier-free guidance, \cite{ho2022classifierfreediffusionguidance}).

\begin{table}[ht]
\centering
\small
\caption{\textbf{VLA adaptation recipes.} Both domains use the same eight-token prefix-conditioning mechanism, but start from different pretrained checkpoints and use independently tuned LoRA and optimization configurations.}
\label{tab:vla_adaptation_recipe}
\resizebox{\columnwidth}{!}{
\begin{tabular}{lcc}
\toprule
\textbf{Configuration} & \textbf{LIBERO} & \textbf{SimplerEnv / Bridge} \\
\midrule
Pretrained checkpoint
& \texttt{lerobot/pi05\_libero\_finetuned\_v044}
& \texttt{juexzz/INTACT-pi0-finetune-bridge} \\
Base VLA
& \(\pi_{0.5}\)
& \(\pi_0\) \\
Base parameters
& Frozen
& Frozen \\
Plan tokens \(k\)
& 8
& 8 \\
Prefix width
& 2048
& 2048 \\
Projector
& \multicolumn{2}{c}{\(d_z\!\rightarrow\!512\!\rightarrow\!512\!\rightarrow\!8\times2048\), SiLU} \\
LoRA target
& Default PEFT target set
& Action-expert attention/MLP projections \\
LoRA rank
& 16
& 16 \\
LoRA \(\alpha\)
& 32
& 16 \\
Projector learning rate
& \(2\times10^{-4}\)
& \(3\times10^{-4}\) \\
LoRA learning rate
& \(1\times10^{-4}\)
& \(2\times10^{-4}\) \\
Batch size
& 16
& 8 \\
Plan dropout \(p_{\mathrm{drop}}\)
& 0.15
& 0.40 \\
Training steps
& 6000
& 4000 \\
Optimizer
& AdamW
& AdamW \\
Training objective
& \multicolumn{2}{c}{Original flow-matching loss of the base VLA} \\
Training seeds
& 6 per demonstration set
& 1 \\
\bottomrule
\end{tabular}}
\end{table}

\subsubsection{Variations of the \pizero{} Baseline}
\label{app:noplan}

When evaluating GPL against \pizero{}, we compare it against one of three distinct baseline configurations: 
\begin{itemize}
    \item \emph{Stock \pizero{}}: The unmodified base model.
    \item \emph{Matched plan-less control}: The GPL model evaluated with the plan tokens omitted.
    \item \emph{Dedicated no-plan adapter}: The stock \pizero{} model fine-tuned for the target task via LoRA, without plan generation.
\end{itemize}

Crucially, these configurations are not interchangeable. As demonstrated on the two-cube grasp task (\Cref{tab:noplan}), simply dropping the plan from GPL (the matched plan-less control) severely degrades performance and does not recover the behavior of the stock model. To ensure a rigorous evaluation, our experiments consistently compare GPL against the strongest of these baselines for each specific test.

\begin{table}[t]
\centering
\caption{
\textbf{Omitting plan tokens does not recover stock \pizero{}.}
Performance comparison on the two-cube grasp task. Relative $L_2$ and cosine similarities are measured against stock \pizero{}. 
}
\label{tab:noplan}
\small
\begin{tabular}{@{}lccc@{}}
\toprule
Model Variant & Rel.\ $L_2$ & Cosine & Grasp Success \\
\midrule
Stock \pizero{} & $0.000$ & $1.000$ & $5/48 = 0.104 \, [0.045, 0.222]$ \\
Matched plan-less & $1.395$ & $0.348$ & $\mathbf{0/48 = 0.000} \, [0.000, 0.074]$ \\
Dedicated no-plan & $1.474$ & $0.283$ & $6/48 = 0.125 \, [0.059, 0.247]$ \\
\bottomrule
\end{tabular}
\end{table}

\subsection{LaRA-Style Matched Baseline}
\label{app:lara}

We implement a LaRA-style matched baseline with a deterministic continuous reasoning latent followed by a generative flow action head, matched to our own conditioning, action-head shape, and evaluation task rather than reproducing LaRA-VLA's original benchmark end to end. Given conditioning features \(c\), the reasoning module produces a single latent \(h_r\) through a two-layer MLP and additionally predicts a visual latent using the auxiliary visual-CoT objective. The action module is a rectified-flow network conditioned on the resulting deterministic reasoning representation \(\hat h\), and is sampled with four Euler steps. We preserve LaRA-VLA's core architectural claim, a deterministic, Coconut-style continuous thought token conditioning a generative action head, trained with its reported three-stage curriculum and visual-CoT auxiliary objective, and adapt the reasoning dimensionality, task domain, and training budget to this paper's own bimodal manipulation substrate and evaluation metrics.

Training follows a three-stage curriculum:
\[
\begin{aligned}
\mathcal{L}_1 &= \mathcal{L}_{\mathrm{plan}} + 0.1\,\mathcal{L}_{\mathrm{vis}},\\
\mathcal{L}_2 &= 0.3\,\mathcal{L}_{\mathrm{plan}}
+0.2\,\mathcal{L}_{\mathrm{vis}}
+\mathcal{L}_{\mathrm{flow}},\\
\mathcal{L}_3 &= \mathcal{L}_{\mathrm{flow}}
+0.05\,\mathcal{L}_{\mathrm{vis}},
\end{aligned}
\]
where \(\mathcal{L}_{\mathrm{plan}}\) is MSE on the deterministic reasoning target, \(\mathcal{L}_{\mathrm{vis}}\) is an \(L_1\) visual-latent prediction loss against an exponential-moving-average (EMA) target encoder (\(\tau=0.99\)), and \(\mathcal{L}_{\mathrm{flow}}\) is the same rectified-flow velocity-matching objective used for GPL. \Cref{tab:lara_recipe} lists the full recipe.

\begin{table}[ht]
\centering
\small
\caption{\textbf{LaRA-VLA baseline recipe.} Architecture and optimization settings of the matched deterministic-latent baseline described in \Cref{app:lara}.}
\label{tab:lara_recipe}
\begin{tabular}{ll}
\toprule
\textbf{Configuration} & \textbf{Value} \\
\midrule
Reasoning latent & Deterministic, \(d_h=3\) \\
Reasoner & 2-layer MLP, SiLU \\
Visual target encoder & EMA, \(\tau=0.99\) \\
Action head & Conditional rectified flow \\
Flow sampling steps & 4 Euler steps \\
Optimizer & Adam \\
Learning rate & \(2\times10^{-3}\) \\
Batch size & 128 \\
Steps per stage & 1500 \\
Training stages & 3 \\
Evaluation seeds & 5 \\
Coverage samples & \(K=16\) per condition \\
\bottomrule
\end{tabular}
\end{table}

For mode-coverage evaluation, we sample \(K=16\) actions for each held-out condition and count a mode as recovered when a committed sample falls within \(0.03\) of the corresponding target. For reasoning-level steering, we apply the same guidance classifier and protocol used for GPL directly to LaRA's deterministic reasoning latent, isolating the effect of replacing a sampled plan distribution with a single continuous thought representation.

\subsection{Learned Plan Coordinates}
\label{app:learned_coords}

To test whether addressability depends on hand-designed plan coordinates, we
learn a plan representation directly from demonstrated end-effector motion.
For each demonstration, we sample the End-Effector Position (EEP) at 16 uniformly spaced
future phases, express each position relative to the current EEP, and flatten
the resulting trajectory to \(x \in \mathbb{R}^{48}\). No object identity, mode
label, or language supervision is used to construct this representation.

We fit a deterministic autoencoder
\[
48 \rightarrow 128 \rightarrow 2 \rightarrow 128 \rightarrow 48
\]
with SiLU activations and reconstruction MSE. Raw trajectories are normalized
using the mean and population standard deviation of the corresponding
outer-training fold; the resulting
2-D latent codes are standardized again using statistics computed only from the same
training fold. The encoder is then frozen, and these standardized codes replace
the hand-designed plan coordinates in the same generative-flow and
deterministic-plan models used in the main experiments. \Cref{tab:learned_coordinate_recipe}
lists the full training configuration.

\begin{table}[ht]
\centering
\small
\caption{\textbf{Learned-coordinate recipe.} The representation is learned
solely by reconstructing future end-effector motion.}
\label{tab:learned_coordinate_recipe}
\begin{tabular}{ll}
\toprule
\textbf{Configuration} & \textbf{Value} \\
\midrule
Trajectory input & 16 future EEP positions \\
Reference frame & Relative to current EEP \\
Input dimension & 48 \\
Encoder & \(48 \rightarrow 128 \rightarrow 2\) \\
Decoder & \(2 \rightarrow 128 \rightarrow 48\) \\
Activation & SiLU \\
Objective & Reconstruction MSE \\
Optimizer & Adam \\
Learning rate & \(2\times10^{-3}\) \\
Autoencoder steps & 800 \\
Latent dimension & 2 \\
Training demonstrations & 400 \\
Base layouts & 40 \\
Jitters per layout & 5 \\
Outer evaluation & 5-fold CV, 32/8 layouts \\
Representation seeds & \(\{0, 1, 2, 3, 4\}\) \\
Coverage samples & \(K = 16\) \\
Validity radius & \(0.05\,\mathrm{m}\) \\
\bottomrule
\end{tabular}
\end{table}

\section{Evaluating Generality: The Maze Navigation Domain}
\label{app:maze}

While the primary focus of this work is robotic manipulation, we also evaluate GPL on a 2D maze navigation task to demonstrate its generality. 

 We build upon the easy-to-hard maze framework of \citet{schwarzschild2021can}. The architecture processes a 3-channel maze image alongside the agent's current cell coordinate. Following our standard recipe, the model first diffuses a \emph{plan} (represented as a 144-dimensional downsampled solution-path image for size-9 mazes) and then autoregresses a sequence of discrete moves conditioned on this generated latent plan. 

Because the original dataset features only single-solution mazes, we developed a custom data generator to create environments with multiple equal-cost, interior-disjoint routes (e.g., a ring, a staircase of two non-crossing paths, and a comb of $K>2$ corridors). This ensures that the target plan distribution is genuinely multimodal at decision forks, and that training and testing sets are strictly disjoint. We train three model variants on identical data: our generative plan model, a standard no-plan baseline, and a deterministic twin that regresses the plan using squared error combined with action cross-entropy.

\begin{figure}[ht]
  \centering
  \includegraphics[width=\linewidth]{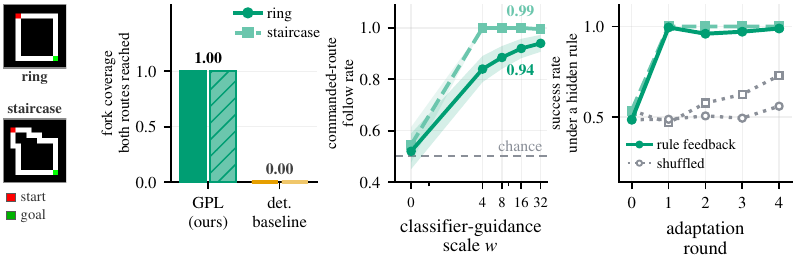}
  \caption{
  \textbf{The three capabilities reproduce in navigation.}
  From left: representative ring and staircase mazes; \(K=16\) samples recover
  both valid routes while the deterministic twin collapses; commanded-route fidelity rises continuously with
  guidance strength; and feedback about a hidden forbidden route raises success
  to \(1.00\) while shuffled feedback lags. Throughout, circle markers are the
  ring maze and square markers the staircase maze, as keyed in the leftmost
  panel.}
  \label{fig:maze}
\end{figure}

We evaluate these models on held-out mazes across three key capabilities, demonstrating the distinct advantages of the generative approach (\Cref{fig:maze}). 

\subsection{Maze-Domain Implementation Details}
\label{app:maze_impl}

The maze experiments use a separate architecture from the manipulation
experiments. All variants share a Vision Transformer (ViT)-style maze encoder and an autoregressive
GPT-2 move decoder over the four discrete actions
\{up, down, left, right\}. The encoder has width 256, depth 6, head dimension
64, RoPE positional encoding, and no dropout. The generative and deterministic
plan models additionally predict the same 144-D plan representation obtained
by bilinearly downsampling the solution-path image by a factor of two,
flattening it, and applying \(L_2\) normalization. The no-plan baseline bypasses
this representation and predicts moves directly from the shared backbone.
\Cref{tab:maze_recipe} gives the full recipe.

\begin{table}[ht]
\centering
\small
\caption{\textbf{Maze-domain training recipe.} All variants share the same
backbone, data, optimizer family, and autoregressive move decoder; only the
plan interface differs.}
\label{tab:maze_recipe}
\resizebox{\columnwidth}{!}{
\begin{tabular}{lccc}
\toprule
\textbf{Configuration}
& \textbf{\GPL{} (generative)}
& \textbf{Deterministic twin}
& \textbf{No-plan baseline} \\
\midrule
Plan representation
& 144-D
& 144-D
& -- \\

Plan head
& Continuous-time diffusion
& Single-pass regression
& None \\

Plan loss
& Diffusion, \(\lambda=5.0\)
& MSE
& -- \\

Action loss
& \multicolumn{3}{c}{Autoregressive move cross-entropy} \\

Optimizer
& \multicolumn{3}{c}{AdamW, weight decay \(0.01\), grad. clip \(1.0\)} \\

Learning rate
& \(5\times10^{-4}\)
& \(5\times10^{-4}\)
& \(3\times10^{-4}\) \\

Batch size
& \multicolumn{3}{c}{64} \\

Warmup
& \multicolumn{3}{c}{200 steps} \\

Precision
& \multicolumn{3}{c}{bfloat16} \\

Training steps
& \multicolumn{3}{c}{6000 (ring) / 4500 (staircase)} \\
\bottomrule
\end{tabular}}
\end{table}

\paragraph{Generative plan model.}
The maze \GPL{} instantiation uses continuous-time diffusion rather than
rectified flow. We use a cosine log-SNR schedule with
\(\gamma_{\min}=-15\) and \(\gamma_{\max}=15\). Training times
\(t\in[0,1]\) are importance-sampled using an EMA loss-weighted sampler over
100 log-SNR bins. At inference, plans are generated with DDIM-style \citep{song2020denoising} sampling
using 50 denoising steps; this value is fixed explicitly by all evaluation
scripts.

\paragraph{Dataset and action horizon.}
We generate 4000 size-9 mazes for each topology with strictly disjoint
training and test sets. Ring mazes contain two equal-length routes around a
rectangular ring, while staircase mazes contain two interior-disjoint winding
routes. The default autoregressive horizon is \(T=8\) moves. Staircase
experiments use \(T=16\), since the longer routes cannot be executed reliably
within eight moves; this keeps closed-loop task completion from confounding the
plan-coverage comparison.

\paragraph{Steering and outcome adaptation.}
Steering uses a noise-conditioned route classifier
\(p(y\mid z_t,\alpha^2)\), trained for 3000 steps, whose log-probability
gradient is added during diffusion sampling. Guidance strength is swept over
\(s\in\{0,4,8,16,32\}\). Outcome adaptation keeps the diffusion generator
frozen and fits a binary success critic to accumulated
\((z,r)\) pairs collected from execution. Candidate plans are then reranked
with Boltzmann weights proportional to
\(\exp(\beta\,g(z))\).

\paragraph{Evaluation protocol.}
Coverage draws \(K=16\) plans at each fork and evaluates closed-loop replanning
on 200 held-out mazes. Steering is evaluated on 100 held-out mazes over the
guidance sweep above. Outcome adaptation uses the hidden route-reliability
intervention and 100 held-out test mazes. Reported confidence intervals capture
variation across held-out mazes; each model variant is trained from a single
training seed per maze family.

\section{Separating Mode Averaging from Addressability}
\label{app:addressable}

The deterministic MSE baseline confounds two effects: it can regress toward an
invalid conditional mean, and at inference time it exposes only a single plan.
We therefore construct a stronger deterministic model that explicitly learns
multiple valid modes and use it to isolate the second effect. Crucially, we
compare two inference read-outs of the \emph{same trained model}: one exposes
only a single mode, while the other preserves access to the learned
alternatives.

\paragraph{Multi-hypothesis deterministic model.}
Given conditioning \(c\), the model predicts \(M=4\) candidate plan means
\(\{\mu_m(c)\}_{m=1}^M\) together with categorical logits over the candidates.
It is trained with WTA multiple-choice regression
\citep{guzmanrivera2012multiple,Rupprecht_2017_ICCV}:
\begin{equation}
    \mathcal{L}
    =
    \frac1B\sum_b
    \min_m
    \left\|\mu_m(c_b)-z_b\right\|_2^2
    +
    \mathrm{CE}\!\left(
        \mathrm{logits}(c_b),
        \arg\min_m
        \left\|\mu_m(c_b)-z_b\right\|_2^2
    \right).
\end{equation}
The first term lets different candidates specialize to different valid plans,
while the second learns their categorical probabilities. We use \(M=4\),
deliberately larger than the two ground-truth modes, so the model is not given
the correct mode count. Conditioning encoder, trunk, optimizer, learning rate,
batch size, and training budget match the deterministic twin.

\paragraph{Same weights, different read-outs.}
After training, we evaluate two inference interfaces without updating any
weights:
\begin{itemize}
    \item \textbf{WTA, MAP:} return only the mean of the highest-probability
    component. The model may have learned several valid modes internally, but
    only one is accessible at inference time.
    \item \textbf{WTA, component sampled:} sample a component index from the
    predicted categorical distribution and return its mean. The same learned
    modes are now individually accessible across repeated draws.
\end{itemize}
This comparison therefore changes \emph{only} whether the alternatives already
encoded in the model remain addressable at inference time.

For steering, the sampled read-out can be commanded by selecting the desired
component, whereas the MAP read-out has no corresponding degree of freedom.
For reranking, we report \emph{rerank feasibility}: the fraction of layouts for
which the \(K=16\) candidate set contains at least one zero-cost plan under an
oracle reranker. This measures whether a useful alternative is present in the
candidate set, independently of critic accuracy.

\begin{table}[ht]
\centering
\caption{
\textbf{Addressability depends on the inference read-out, not only on whether
the model has learned multiple modes.}
Both rows use the \emph{same WTA-trained weights}; only the read-out changes.
Eight held-out layouts, five seeds, \(K=16\). Intervals are Student-\(t\)
intervals over seeds, truncated to the support of each metric.
}
\label{tab:addressable}
\small
\setlength{\tabcolsep}{4pt}
\resizebox{\linewidth}{!}{%
\begin{tabular}{lcccc}
\toprule
read-out
& invalid draw
& P(cover all)
& steer range
& rerank feasible \\
\midrule
WTA, MAP
& $0.03\,[0.00,0.09]$
& $0.00$
& $0.00$
& $0.50\,[0.39,0.61]$ \\

WTA, component sampled
& $0.07\,[0.05,0.09]$
& $0.90\,[0.83,0.97]$
& $+0.48\,[+0.46,+0.51]$
& $1.00$ \\
\bottomrule
\end{tabular}}
\end{table}

The result separates mode averaging from addressability
(\Cref{tab:addressable}). The WTA model already avoids the invalid conditional
mean by learning multiple candidate plans. Nevertheless, MAP inference still
collapses those learned alternatives to a single exposed plan, giving zero
coverage and zero steering range. Changing only the read-out to component
sampling recovers coverage, steerability, and full reranking feasibility.
Thus, avoiding mode averaging is not sufficient: the learned alternatives must
also remain accessible at inference time.

We therefore treat the WTA experiment as a positive control for the
addressability hypothesis rather than as a competing method. It shows that the
same property can arise from a discrete mixture representation: rectified flow
is not uniquely required. \GPL{} instead provides a continuous plan
distribution without requiring a predefined component count \(M\), while
supporting the same sampling, steering, and outcome-reweighting interface.

\section{Steering: Where the Handle Acts}
\label{app:group_steering}
Supports the steering results of \Cref{subsec:steer}: where guidance is most effective under a shared executor (\Cref{app:steering_locus}), the protocol behind the native-\pizero{} action-chunk comparison (\Cref{app:action_guidance}), and the full language-steering baseline comparison (\Cref{app:language_steering}).

\subsection{Full Steering-Locus Comparison}
\label{app:steering_locus}

\begin{figure}[ht]
  \centering
  \includegraphics[width=\linewidth]{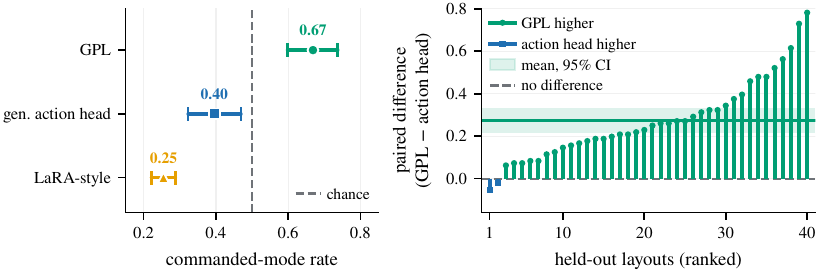}
  \caption{
  \textbf{Steering is stronger at the plan locus under a shared executor.}
  \emph{Left:} Commanded-mode rate at the pre-fixed operating point $w=16$
  for the \GPL{} plan latent, the generative action head, and the LaRA-style
  deterministic reasoning token. Intervals are $95\%$ bootstraps clustered
  over the $40$ held-out layouts.
  \emph{Right:} Per-layout paired difference, ranked
  over the same layouts. The plan locus is favored on $38/40$ layouts.
  }
  \label{fig:steer_locus_main}
\end{figure}

First, we isolate the locus of intervention using a shared soft-token executor (a flow-matching action model trained from scratch). To ensure a fair comparison, the plan and action branches share the exact same task, layouts, and guidance objective; the only variable is the representation being guided. We additionally compare against a LaRA-style baseline. Although its downstream action head is generative, its deterministic reasoning token yields a steer range of nearly zero. As \Cref{fig:steer_locus_main} demonstrates, the plan representation serves as a substantially stronger locus for steering.

\subsection{Guidance on the Native High-Dimensional Action Chunk}
\label{app:action_guidance}

This section gives the protocol behind the main-text comparison in
\Cref{fig:action_guidance_native}: how guidance doses are made comparable
across representations and how discriminator capacity is matched between
the plan and action-chunk arms.

The main-text comparison with \pizero{} evaluates each intervention through
its native executor. The plan arm uses the plan-conditioned adapter, whereas
the action arm uses a dedicated no-plan adapter trained at the same adaptation
budget. Each arm receives its own discriminator: plan guidance differentiates
with respect to the noisy plan $\zr_t$, while action guidance trains the
requested-mode classifier on that adapter's noisy native action chunk and
differentiates with respect to
$x_t\in\mathbb{R}^{4\times32}$.

This is therefore a matched-system comparison rather than a causal isolation
of locus alone. As a diagnostic, the commanded cube is explicit in the 3-D
plan, whereas held-out balanced accuracy for decoding it from the action chunk
is $0.57$, $0.57$, and $0.49$ across the three train/evaluation pairings.
Thus the comparison measures the effectiveness of the two native intervention
interfaces; it does not establish that representation dimensionality alone
causes the difference.

\paragraph{Making the guidance doses comparable.}
A raw classifier-gradient magnitude is not directly comparable across
representations with different dimensionality and scale. We therefore
normalize each guidance displacement by the corresponding Euler-step length:
\begin{equation}
    x_{t+\Delta t}
    =
    x_t
    +
    \Delta t\,v_t
    +
    w\,
    \left\|\Delta t\,v_t\right\|\,
    \frac{
      \nabla_{x_t}\log q_\psi(y\mid x_t,t)
    }{
      \left\|
      \nabla_{x_t}\log q_\psi(y\mid x_t,t)
      \right\|
    }.
\end{equation}
With this parameterization, $w$ is dimensionless and $w=0$ recovers the
unguided sampler in both arms.

\subsection{Language Steering Under a Shared Executor}
\label{app:language_steering}

We next compare the latent plan against natural language as an alternative command channel. To ensure any differences are due to the command channel and not downstream execution capability, we train a single shared executor per seed on the two-cube task. This executor exposes three independently maskable command channels. During evaluation, each arm activates only its own channel and nulls the others, and all arms are scored using the same closed-loop outcome.

We compare the plan against three language baselines:
\begin{itemize}
    \item \textbf{L-stock:} A frozen ECoT model decodes a cube choice and routes that decision to the shared downstream executor. 
    \item \textbf{L-prefix:} The named instruction is provided directly as text in the shared executor's own LM prefix, with no separate reasoner or learned projector.
    \item \textbf{L-soft:} Projects a $4096$-D sentence feature into the exact same eight-soft-token interface used by the plan. This has 12.27\% more trainable parameters than the \GPL{} projector, ensuring the plan does not win simply due to parameter capacity.
\end{itemize}

\begin{table}[ht]
\centering
\caption{
\textbf{Language and plan steering under a shared executor.}
All arms drive the same executor and are scored identically.
Intervals are $95\%$ bootstraps clustered over the eight held-out layouts;
grasp rate is averaged over seeds.
}
\label{tab:lang_arms}
\small
\setlength{\tabcolsep}{3.5pt}
\resizebox{\linewidth}{!}{%
\begin{tabular}{llcccc}
\toprule
arm
& command interface
& seed 0
& seed 1
& seed 2
& grasp \\
\midrule

\GPL{}
& plan latent
& $+0.70\,[+0.47,+0.90]$
& $+0.48\,[+0.23,+0.73]$
& $+0.45\,[+0.18,+0.75]$
& $0.33$ \\

L-soft
& sentence feature
& $+0.00\,[0.00,0.00]$
& $-0.08\,[-0.18,0.00]$
& $+0.00\,[0.00,0.00]$
& $0.06$ \\

L-prefix
& instruction
& $+0.13\,[-0.05,+0.35]$
& $+0.20\,[+0.03,+0.45]$
& $+0.20\,[+0.03,+0.43]$
& $0.13$ \\

L-stock
& ECoT
& $-0.05\,[-0.30,+0.13]$
& $+0.03\,[-0.05,+0.10]$
& $+0.08\,[+0.03,+0.15]$
& $0.50$ \\

\midrule
\GPL{} $-$ L-stock
& paired difference
& $+0.75\,[+0.40,+1.13]$
& $+0.45\,[+0.20,+0.75]$
& $+0.38\,[+0.13,+0.60]$
& -- \\

\bottomrule
\end{tabular}}
\end{table}

\textbf{Execution versus steering.} L-stock provides the most interpretable language baseline because execution competence does not limit it. It achieves the highest overall grasp rate ($0.50$), yet its steer-range interval contains zero on two of the three seeds (\Cref{tab:lang_arms}). This indicates that while L-stock is physically capable of the task, it largely ignores the language command. In contrast, the plan (\GPL{}) successfully steers execution. 

\textbf{L-soft failure.} While L-soft provides a fair architectural control, its grasp rate collapsed to $0.06$ at this data scale. Because it rarely grasps anything, its steer range cannot serve as a reliable measure of information transfer. 

\textbf{Guidance dose.} For the language channels evaluated here, preliminary dose calibration showed only a narrow useful extrapolation range (peaking near $w=0.5$ and collapsing by $w=1.0$). Therefore, we report the plain conditional performance for each language channel rather than selecting a guidance dose on the evaluation set.

\section{Outcome Adaptation and the Compute Frontier}
\label{app:group_adaptation}

Supports the outcome-adaptation results of \Cref{subsec:si}: the reranking
protocol, its cost validation, where reranking should act, and the
compute-frontier measurement. The frontier is measured on the controlled
two-cube task rather than through \pizero{}, and we do not extrapolate its ratio
to the \pizero{} experiments.

\subsection{Additional Outcome Adaptation Details}
\label{app:si_details}

This section formalizes the outcome-adaptation mechanism sketched in
\Cref{sec:method}: the reweighted-distribution estimator and the
scoring functions used across the outcome-adaptation experiments.

The main text treats outcome adaptation as the general operation of reweighting
samples from a frozen plan distribution using deployment feedback.
Over the finite candidate set, the reweighted distribution of
\Cref{eq:si_distribution} is implemented as
\begin{equation}
    \rho_k
    =
    \frac{
        \exp\!\left(\beta s(\Ccond,\zr^{(k)})\right)
    }{
        \sum_{j=1}^{K}
        \exp\!\left(\beta s(\Ccond,\zr^{(j)})\right)
    },
    \qquad
    \zr^\star
    \sim
    \operatorname{Categorical}(\rho_1,\ldots,\rho_K).
\end{equation}

\textbf{Reporting convention.} Instead of executing a single sampled plan $\zr^\star$, all results in this paper report the Boltzmann-weighted \emph{expected} outcome evaluated exactly over the sampled pool:
\begin{equation}
    \hat V(\Ccond) = \mathbb{E}_{\zr^\star\sim\operatorname{Categorical}(\rho)}[r^\star] = \sum_{k=1}^{K} \rho_k\, r^{(k)}.
    \label{eq:si_expectation}
\end{equation}
Here, $r^{(k)}$ is the observed outcome (e.g., a success bit, path cost, or preference indicator) associated with candidate $\zr^{(k)}$. This approach applies uniformly to all three scoring methods below. It provides a lower-variance estimate than a single physical draw, though it requires knowing the actual outcome for all $K$ candidates rather than just the one deployed.

\paragraph{Learned critic.}
We fit a scalar regressor \(g_\omega:\mathbb{R}^{d_z}\to\mathbb{R}\) to the
accumulated \((\zr^{(k)},r^{(k)})\) pairs observed at deployment, using a
supervised loss matched to the outcome type (binary cross-entropy for a
success bit, squared error for a scalar cost), and set
\(s(\Ccond,\zr)=g_\omega(\zr)\). Because \(g_\omega\) is continuous in
\(\zr\), it can score plans that were never directly observed, at the cost
of the fitting compute.

\paragraph{Reward table.}
When the outcome space admits a natural finite partition of the plan into
\(M\) discrete modes through a mode map \(m:\mathbb{R}^{d_z}\to\{1,\ldots,M\}\)
(e.g., which object, route, or order a candidate plan resolves to), the
score can instead be read directly off observed outcomes in that mode:
\begin{equation}
    s(\Ccond,\zr) = \bar r_{m(\zr)}, \qquad
    \bar r_j = \frac{\sum_{i=1}^{N} \mathbb{1}[m(\zr^{(i)})=j]\, r^{(i)} + \kappa}
                     {n_j + 2\kappa},
    \label{eq:reward_table}
\end{equation}
where \(n_j\) is the number of observed rollouts assigned to mode \(j\) and
\(\kappa>0\) is a Jeffreys/Laplace pseudo-count keeping \(\bar r_j\) finite
before mode \(j\) has been observed (\(\kappa=0.5\) throughout). This score
costs zero training FLOPs, since it is a running average rather than a
fitted function, but it can only distinguish plans across the modes the
partition resolves.

\subsection{Where Should Outcome-Based Reranking Act?}
\label{app:rerank_locus}

To rigorously isolate the locus of reranking, we compare the plan arm against a \emph{dedicated no-plan adapter} (\Cref{app:noplan}), matching the architecture, optimizer, rank, and step budget to avoid the degradation of a network simply run without its plan. Both LoRA adapters are hot-swapped in a single process, sharing identical scenes, backbones, execution trajectories, and per-candidate action-noise seeds.

\begin{figure}[ht]
  \centering
  \includegraphics[width=\linewidth]{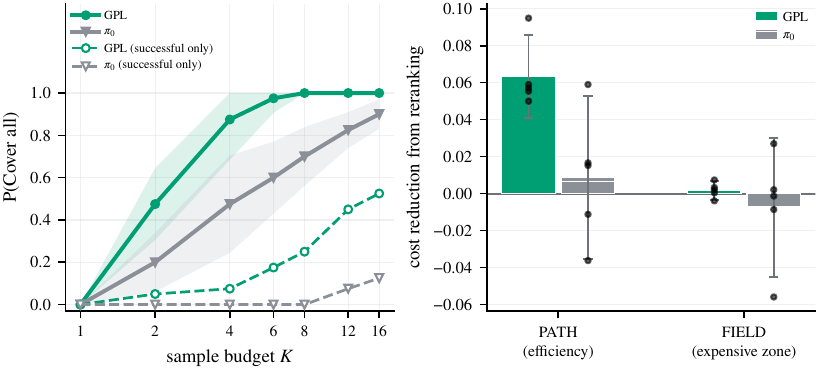}
  \caption{
  \textbf{Candidate coverage and reranking cost reduction.}
  \textbf{Left:} Coverage vs. sample budget \(K\). Solid curves denote any committed draw; dashed curves isolate successful grasps.
  \textbf{Right:} Cost reduction at \(K=6\) for both the PATH (efficiency) and FIELD (expensive zone) objectives.
  \textbf{Setup:} 5 seeds \(\times\) 8 held-out scenes. Critic capacity ($4{,}481$ parameters), updates ($700$), samples per update ($64$), and noise seeds are matched across arms. Overall grasp success is \(0.295\) (plan) vs. \(0.133\) (action).
  }
  \label{fig:rerank_locus}
\end{figure}

As discussed in the main text, the performance gap between plan- and action-level reranking stems from executable diversity (\Cref{fig:rerank_locus}, left). While the action arm achieves comparable raw candidate coverage at higher budgets, its draws successfully grasp at less than half the rate of the plan arm. Because the action sampler rarely produces independent, successful alternatives, the downstream reranking outcomes differ significantly depending on the objective (\Cref{fig:rerank_locus}, right). For the PATH objective (total end-effector Euclidean distance), plan-level reranking significantly reduces execution cost while action-level reranking does not, and this advantage widens with the sample budget \(K\). Conversely, both arms perform similarly on the FIELD objective (a smooth Gaussian cost bump centered on one of the two cube locations in the scene), where cost relies on localized spatial avoidance rather than being tightly coupled to the initial mode choice.

\subsection{Compute-Frontier Protocol}
\label{app:frontier}

This section specifies the success--training-FLOPs comparison in
\Cref{subsec:si}: two frozen-\GPL{} outcome-adaptation variants, continued
behavior cloning, and AWR \citep{peng2019awr}.

\paragraph{Task and evaluation.}
We use the two-cube reliability task, in which one grasp mode succeeds with
probability \(0.9\) and the other with probability \(0.25\). These
reliabilities are hidden from the policy; after each executed grasp, the
environment returns only a binary success/failure outcome. All arms start from
the same conditioning cache, plan-latent flow, deterministic twin, and
flow-action executor, and differ only in how deployment outcomes are used.

At every round, performance is measured as the Boltzmann-weighted expected
success rate over the sampled plan pool (\Cref{eq:si_expectation}). We evaluate
two forms of outcome adaptation. \textbf{\GPL{} + critic} learns a score model
from accumulated deployment outcomes, while \textbf{\GPL{} + reward table}
directly estimates the empirical success probability of each discrete mode.
The latter is applicable here because the two plan modes correspond directly
to the left and right cube. It reaches a final \SR{} statistically
indistinguishable from the learned critic while requiring no score-model
training (\Cref{fig:costsi}B).

\paragraph{Compared arms.}
The four strategies are summarized in \Cref{tab:frontier_arms}. Continued BC and AWR update their flow cumulatively across five deployment rounds, with no reset between rounds. For every arm, we additionally evaluate a shuffled-outcome control in which observed rewards are randomly reassigned to sampled plans. These controls test whether improvement depends on the
correspondence between a plan and its observed outcome rather than arising from reranking or continued optimization alone. 

\begin{table}[ht]
\centering
\small
\caption{\textbf{Compute-frontier arms.} Each deployment round uses \(K=16\) sampled plans per training scene. Training costs reflect the computational budget per round.}
\label{tab:frontier_arms}
\resizebox{\columnwidth}{!}{
\begin{tabular}{>{\raggedright\arraybackslash}p{0.18\linewidth} >{\raggedright\arraybackslash}p{0.22\linewidth} >{\raggedright\arraybackslash}p{0.28\linewidth} >{\raggedright\arraybackslash}p{0.21\linewidth}}
\toprule
\textbf{Arm} &
\textbf{Outcome data retained} &
\textbf{Update rule} &
\textbf{Training cost} \\
\midrule

\textbf{\GPL{} + critic} &
All observed \((\zr,r)\) pairs &
Fit a \(\sim17\)k-parameter critic on accumulated outcomes;
reweight sampled plans with the critic score using
\Cref{eq:si_expectation}. &
Critic training only; no generator updates. \\

\textbf{\GPL{} + reward table} &
Per-mode successes and trials &
Estimate each mode's empirical success probability; use this
score in the same Boltzmann reweighting rule
(\Cref{eq:reward_table}). &
Zero training FLOPs. \\

\textbf{Continued BC} &
Successful samples only (\(r=1\)), accumulated across rounds &
Fine-tune the flow on the accumulated successful-plan pool using the original
flow-matching objective. &
1500 Adam steps/round, \(2\times10^{-3}\) learning rate, batch size 128. \\

\textbf{AWR} &
All samples, successes and failures, accumulated across rounds &
Compute \(A=r-\bar r\), resample examples with probability
\(\propto\exp(\tau_{\mathrm{AWR}} A)\), \(\tau_{\mathrm{AWR}}=4\), and fine-tune with the original
flow-matching objective. &
1500 Adam steps/round, \(2\times10^{-3}\) learning rate, batch size 128. \\

\bottomrule
\end{tabular}}
\end{table}

\paragraph{FLOP accounting.}
Training cost is estimated with the standard proxy:
\[
\mathrm{FLOPs}
=
6\,N_{\mathrm{params}}\,N_{\mathrm{samples\ processed}},
\]
and accumulated over all optimizer steps and deployment rounds.
\textbf{\GPL{} + reward table} incurs exactly zero training FLOPs, while
\textbf{\GPL{} + critic} pays only for fitting the
\(\sim17\)k-parameter critic. Continued BC and AWR
repeatedly update the larger flow and therefore each consume \(388\times\)
the training FLOPs of the critic-based \GPL{} arm, yielding the ratio reported
in \Cref{fig:costsi}B.

Because the models in this controlled experiment are deliberately small, we
interpret this comparison as a \emph{relative compute frontier} under a common
FLOP proxy rather than an absolute systems benchmark. We do not report
wall-clock time, and the \(388\times\) ratio is specific to this task; we do
not extrapolate it to the \pizero{} experiments.

\section{The Redundancy Score and the LIBERO Counterfactual Test}
\label{app:redundancy-libero}

This section develops and validates the redundancy score $\widetilde{R}$ presented in \Cref{sec:when}.

\subsection{Counterfactual Intervention}
\label{app:libero-endpoints}

We first test whether reducing redundancy makes the plan more influential. The original scene demonstrations partially couple the visual scene to the demonstrated decision. We replace them with counterfactual pairs in which opposing decisions occur in the same visual context. This removes the observation-decision correlation from the plan-training data. The executor, benchmark, adaptation procedure, and six adapter seeds remain unchanged.

\begin{table}[ht]
\centering
\caption{
\textbf{Lower redundancy increases plan use under a matched intervention.}
\emph{Follow} is the fraction of grasped episodes reaching the commanded bowl.
Steer-range intervals are scene-clustered $95\%$ bootstraps. All six seeds are
matched across demonstration sets and favor the counterfactual intervention on
each steering endpoint.
}
\label{tab:libero-endpoints}
\small
\resizebox{\linewidth}{!}{%
\begin{tabular}{lccccc}
\toprule
demonstrations
& \multicolumn{2}{c}{follow}
& steer range, grasped
& steer range, per episode
& grasp cost \\
& cmd 1 & cmd 2 & & & \\
\midrule
habit ($\widetilde{R}=0.47$)
& 0.897
& 0.294
& +0.190 [+0.146,+0.235]
& +0.108 [+0.083,+0.136]
& 0.250 \\

counterfactual ($\widetilde{R}=0$)
& 0.974
& 0.672
& +0.646 [+0.584,+0.708]
& +0.425 [+0.374,+0.478]
& 0.274 \\
\bottomrule
\end{tabular}}
\end{table}

Adding counterfactual scenes makes the plan substantially more relevant to the decision without degrading the overall competence of the policy (\Cref{tab:libero-endpoints}).

\subsection{Estimating Redundancy}
\label{app:redundancy-score}

To ensure our conclusions are not dependent on a specific metric, we estimate the redundancy score using three different methods: 

\begin{itemize}
    \item \textbf{E1:} The preregistered L2 logistic probe with fixed regularization (used for the main text endpoints).
    \item \textbf{E2:} The same probe with regularization selected by inner grouped cross-validation and calibrated probabilities.
    \item \textbf{E3:} A non-parametric grouped five-fold $k$-NN conditional-entropy estimate ($k=25$, cosine distance).
\end{itemize}  

We replace a fraction $q$ of the original scenes (drawn once per scene with common random numbers, so the five levels are nested) with their counterfactual counterparts, creating five matched demonstration mixtures, where $q=0$ denotes the original scenes and $q=1$ the fully counterfactual set.
While the absolute magnitudes vary, all three estimators preserve the exact same redundancy ordering across demonstration sets (\Cref{tab:libero-dose}).

\begin{table}[ht]
\centering
\caption{
\textbf{Redundancy dose response.}
All five mixtures use the same six-task protocol and $360$ episodes per arm.
E1 is the preregistered fixed-regularization probe (used for the main-text
endpoint), E2 is tuned and probability-calibrated (used for this sweep's finer
resolution), and E3 is the grouped $k$-NN conditional-entropy estimate
(\Cref{app:redundancy-score}). Steer-range intervals are scene-clustered
$95\%$ bootstraps.
}
\label{tab:libero-dose}
\small
\resizebox{\linewidth}{!}{%
\begin{tabular}{lcccll}
\toprule
$q$ & E1 & E2 & E3 & steer range, grasped & steer range, per episode \\
\midrule
0.00 & 0.061 & 0.234 & 0.139 & +0.236 [+0.155,+0.326] & +0.181 [+0.119,+0.250] \\
0.25 & 0.051 & 0.110 & 0.024 & +0.444 [+0.335,+0.569] & +0.272 [+0.206,+0.347] \\
0.50 & 0.019 & 0.052 & 0.017 & +0.556 [+0.428,+0.695] & +0.336 [+0.255,+0.426] \\
0.75 & 0.005 & 0.005 & 0.000 & +0.596 [+0.486,+0.710] & +0.408 [+0.325,+0.495] \\
1.00 & 0.000 & 0.000 & 0.000 & +0.640 [+0.523,+0.760] & +0.378 [+0.300,+0.462] \\
\bottomrule
\end{tabular}}
\end{table}

The results in \Cref{tab:libero-dose} demonstrate a continuous relationship: decreasing redundancy correlates perfectly with increased steering (Spearman's $\rho=-1.00$). Methodologically, this dose-response sweep isolates six tasks and samples exactly one record per usable training episode. In contrast, the baseline in \Cref{tab:libero-endpoints} spans ten tasks and retains every logged conditioning vector, including repeated physical attempts at the same scene.

\section{Generality Across Plan Representations and Domains}

\subsection{A Shared 54-D Contact Sketch}
\label{app:contact-sketch}

We evaluate a shared plan representation across four SimplerEnv task families
with qualitatively different decision ambiguities:
\begin{itemize}
    \item \emph{pick-and-place}, where a generic instruction leaves two identical
cubes as valid targets;
\item \emph{grasp-yaw}, where a square-footprint cube admits two valid grasp
orientations;
\item \emph{can pick-or-push}, where the same target can be reached through
carrying or pushing; 
\item \emph{sweep}, a symmetric non-prehensile task in which debris can be pushed
toward either of two identical corners.
\end{itemize}

To represent these decisions through one interface, we use a shared 54-D
contact sketch with six object slots and nine channels per slot: object-relative
contact position, approach direction, and contact state. Unused slots are
padded and overflow objects are truncated by a fixed rule. Different families
therefore express their decision through different components of the same
representation: pick-and-place and grasp-yaw primarily through contact position
and approach direction, can pick-or-push through gripper and surface-contact
state, and sweep through target contact geometry.

\begin{figure}[ht]
  \centering
  \includegraphics[width=0.85\linewidth]{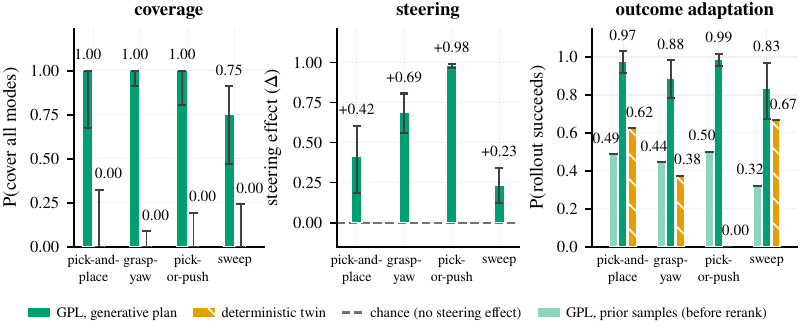}
  \caption{
  \textbf{One shared 54-D contact sketch, one flow reasoner, four SimplerEnv
  families, three capabilities.}
  \emph{Left:} coverage of every valid mode by $K$ plan draws.
  \emph{Middle:} steering effect, measured as commanded-mode rate minus the
  off-diagonal rate. Both panels are scored in plan space, by assigning each
  drawn sketch to its nearest mode centroid; no executor is run.
  \emph{Right:} outcome adaptation under the family-specific physical
  deployment shifts detailed below. Each reward is the success of a scripted
  controller executing that mode in the shifted scene. Bars show the frozen plan prior, the same
  prior reranked from measured rollout outcomes, and the deterministic twin;
  each family also includes an unshifted no-signal control.
  }
  \label{fig:contact_sketch}
\end{figure}

Despite this shared representation, a single jointly trained reasoner preserves addressability across all four families (\Cref{fig:contact_sketch}). 

\paragraph{A strategy-level deployment shift.}
\textit{Can pick-or-push} provides a particularly direct example because the
two modes correspond to different physical strategies. The reasoner is trained
only on small cans (diameters $29.7$--$36.3$\,mm), for which both carrying and
pushing are viable, while deployment changes only the can diameter. Although
the WidowX gripper opens wider than these objects, grasping becomes limited by
the approach geometry: as the can widens, the open fingers collide with it
during the descending pick approach and topple it before the close command.

We verify this mechanism directly. At a $39.6$\,mm diameter, the can remains
upright after the lateral reach in all $20$ instrumented episodes, but crosses
the tipping threshold during the subsequent descent in all $20$, before any
gripper-close command. Pushing avoids this approach and remains viable. The frozen generator nevertheless continues to propose both strategies on
shifted scenes. The deployment outcomes therefore change which existing plan
is preferred, rather than requiring a new strategy to be learned
(\Cref{fig:contact_sketch}).

\paragraph{Scene-dependent shifts across task families.}
We apply a physical intervention that selectively disables one alternative in
each family. Pick-and-place enlarges one cube beyond the usable grasp geometry;
grasp-yaw introduces a barrier along one grasp axis; can pick-or-push uses the
size intervention analyzed above; and sweep places a lip across the approach
to one target corner.

For pick-and-place, grasp-yaw, and sweep, the disabled alternative is randomized
independently across scenes. Consequently, neither mode has a globally higher
marginal reward: successful adaptation requires predicting \emph{which}
alternative remains viable in the current scene. The can shift instead induces
a persistent preference for pushing across deployment scenes. Together, these
settings distinguish scene-conditioned adaptation from learning a single
global mode preference. \Cref{tab:contact_sketch_protocol} reports the per-family scene and
rollout counts.

\begin{table}[ht]
\centering
\small
\caption{\textbf{Contact-sketch battery: per-family scene and rollout counts.}}
\label{tab:contact_sketch_protocol}
\begin{tabular}{lcccc}
\toprule
 & pick-and-place & grasp-yaw & can pick-or-push & sweep \\
\midrule
Training scenes & 320 & 320 & 104 & 78 \\
Held-out (test) scenes & 80 & 80 & 32 & 24 \\
Unique scene configurations & 40 & 200 & 68 & 51 \\
Scenes with a measured rollout outcome & 32 & 160 & 52 & 39 \\
\bottomrule
\end{tabular}
\end{table}

\paragraph{Protocol details.}
The shared evaluation protocol is summarized in
\Cref{tab:contact_sketch_eval_protocol}.

\begin{table}[ht]
\centering
\small
\caption{\textbf{Contact-sketch evaluation protocol.} Sampling and uncertainty
estimation are fixed across all four task families.}
\label{tab:contact_sketch_eval_protocol}
\resizebox{\columnwidth}{!}{
\begin{tabular}{lccc}
\toprule
\textbf{Setting}
& \textbf{Coverage}
& \textbf{Steering}
& \textbf{Outcome adaptation} \\
\midrule
Samples per scene / mode
& \(K=48\)
& \(K=24\)
& \(K=32\) \\

Flow integration
& 4 Euler steps
& 4 Euler steps
& 4 Euler steps \\

Selection / guidance
& Unguided sampling
& Classifier guidance, \(\lambda=6.0\)
& Boltzmann reranking, \(\beta=5.0\) \\

Outcome model
& --
& --
& BCE-trained critic \\

Physical feedback
& --
& --
& One scripted-controller rollout per scene and mode,
shared across its \(K\) candidates \\

Reported quantity
& Mode coverage
& Commanded-mode probability
& Post-minus-prior \(\SR\) gain \\

Uncertainty
& 95\% Wilson interval
& --
& 95\% paired bootstrap interval \\

Evaluation aggregation
& \multicolumn{3}{c}{One held-out evaluation pass; no averaging over training seeds} \\
\bottomrule
\end{tabular}}
\end{table}

\subsection{Cross-Simulator Battery}
\label{app:cross_sim}

\begin{figure}[h]
  \centering
  \includegraphics[width=0.85\linewidth]{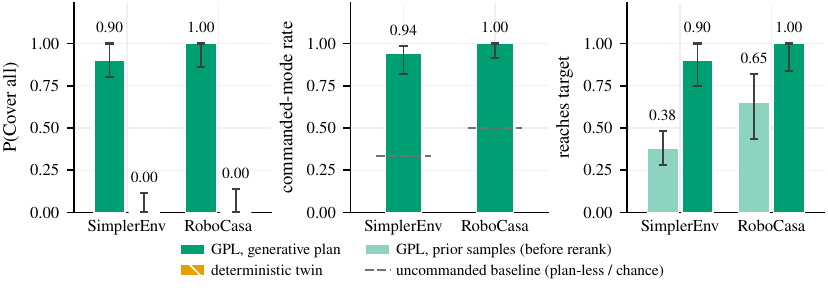}
  \caption{
  \textbf{The three capabilities replicate in a second simulator and
  embodiment.}
  SimplerEnv is closed-loop: coverage and adaptation use a two-spoon scene,
  steering a three-cube scene, scored by the object the gripper approaches.
  RoboCasa is evaluated offline on commanded relative pose, with fixed plan
  anchors in place of a trained reasoner. Terminal task completion
  additionally requires mobile-base navigation, which lies outside the adapted
  executor.
  }
  \label{fig:cross_sim}
\end{figure}

RoboCasa changes the simulator and scene distribution, while retaining the same plan interface
and pretrained \pizero{} executor. \Cref{fig:cross_sim} reports coverage, steering, and outcome
adaptation under this shift.  Coverage there tests that \pizero{} realizes both commanded directions; the
deterministic twin is the midpoint anchor, and the adaptation prior draws anchors uniformly
between the two sides.

\end{document}